\documentclass{article}

\usepackage{microtype}
\usepackage{graphicx}
\usepackage{subfigure}
\usepackage{booktabs} 
\usepackage{amsmath}
\usepackage{amssymb}

\usepackage{hyperref}
\usepackage{cleveref}
\usepackage{siunitx}

\usepackage[accepted]{mlsys2024}

\mlsystitlerunning{\textbf{TREE}: \underline{T}ree \underline{R}egularization for \underline{E}fficient \underline{E}xecution}

\newcommand{\Cpp}{\mbox{C\texttt{++}}}

\begin{document}

\twocolumn[
\mlsystitle{\textbf{TREE}: \underline{T}ree \underline{R}egularization for \underline{E}fficient \underline{E}xecution}

\mlsyssetsymbol{equal}{*}

\begin{mlsysauthorlist}
\mlsysauthor{Lena Schmid}{TU}
\mlsysauthor{Daniel Biebert}{TU}
\mlsysauthor{Christian Hakert}{TU}
\mlsysauthor{Kuan-Hsun Chen}{NL}
\mlsysauthor{Michel Lang}{TU}
\mlsysauthor{Markus Pauly}{TU}
\mlsysauthor{Jian-Jia Chen}{TU}
\end{mlsysauthorlist}

\mlsysaffiliation{TU}{TU Dortmund University, Germany}
\mlsysaffiliation{NL}{University of Twente, Netherlands}

\mlsyscorrespondingauthor{Lena Schmid}{lena.schmid@tu-dortmund.de}

\mlsyskeywords{Machine Learning, MLSys}

\vskip 0.3in

\begin{abstract}
The rise of machine learning methods on heavily resource constrained devices requires not only the choice of a suitable model architecture for the target platform, but also the optimization of the chosen model with regard to execution time consumption for inference in order to optimally utilize the available resources. Random forests and decision trees are shown to be a suitable model for such a scenario, since they are not only heavily tunable towards the total model size, but also offer a high potential for optimizing their executions according to the underlying memory architecture. 

In addition to the straightforward strategy of enforcing shorter paths through decision trees and hence reducing the execution time for inference, hardware-aware implementations can optimize the execution time in an orthogonal manner.
One particular hardware-aware optimization is to layout the memory of decision trees in such a way, that higher probably paths are less likely to be evicted from system caches. This works particularly well when splits within tree nodes are uneven and have a high probability to visit one of the child nodes.

In this paper, we present a method to reduce path lengths by \emph{rewarding} uneven probability distributions during the training of decision trees at the cost of a minimal accuracy degradation. 
Specifically, we regularize the impurity computation of the CART algorithm in order to favor not only low impurity, but also highly asymmetric distributions for the evaluation of split criteria and hence offer a high optimization potential for a memory architecture-aware implementation.
We show that especially for binary classification data sets and data sets with many samples, this form of regularization can lead to an reduction of up to $\approx 4\times$ in the execution time with a minimal accuracy degradation. 
\end{abstract}
]

\printAffiliationsAndNotice{}  

\thispagestyle{plain}

\section{Introduction}
Execution time optimization of machine learning models on the edge on extremely resource constrained devices has been widely studied, especially known as TinyML scenarios.
While one popular approach is to shrink the models (e.g., by reducing the number of neurons in neural networks, or the depth of decision trees in random forests) without losing much accuracy, this approach is agnostic to the actual properties of the underlying hardware. One aspect of resource limitation is often the limited availability of energy and hence time budget for the execution of inference. Shrinking models indeed can meet this requirement, but a considerable reduction of the execution time can also be achieved by an orthogonal hardware-aware implementation of the model, especially in the context of random forests~\cite{chen2022efficient, 9925686}.



Random forests and their inner structure of decision trees are a premier candidate for cache-aware optimizations, since every inference follows one path, 
requiring only a small subset of nodes from the tree.
This naturally fits in the design principle of caches, since these are usually small and depend on a high locality of the memory accesses to be fast and efficient. 
Chen et~al. leverage a probabilistic model, describing the distribution of splits, to place the frequent accessed paths in a cache-friendly manner~\cite{chen2022efficient}. 
Breaking the probabilistic model down to a single tree node, we observe that the approach can be beneficial only when the probability of data tuples (i.e., split) to take the left branch or the right branch differs significantly.
In consequence, when optimizing the execution time of random forest models, a reduction in the tree size should account for the distribution of splits in the tree nodes and maintain or enforce the property of uneven splits. This can lead to a considerable execution time improvement by the reduced tree size and an orthogonal improvement to favor such cache-friendly implementations.

In this paper, 
we introduce the design of a hardware-aware regularization for decision tree training
by actively \emph{rewarding} uneven splits in single decision tree nodes.
This leads to the regularized construction of decision trees, which maintain the crucial properties for cache optimization, but with reduced total size or depth of some paths. 
Consequently, the \textit{studied problem} of this paper is how to regularize random forest training with the objective of reducing the model size and to reward uneven splits, while not degrading the accuracy significantly. 
We \textit{tackle the problem} by introducing a regularization term into the split method of decision trees. This regularizer rewards split decisions that lead to uneven splits to uphold asymmetric distributions.
This leads to an orthogonal speed improvement to the cache optimization mentioned above, and can even assist cache optimization.

The introduction of the regularization offers a trade-off for the application. The regularization term can be controlled by a factor to take an either minor or major influence. We propose an intuitive application, where a tolerable degradation in accuracy can be defined by the user. Subsequently, possible degrees of regularization are automatically tested, and the configuration with the maximal improvement in execution time within the tolerable accuracy degradation is chosen.
If users are eager for a deeper investigation and the manual choice of a trade-off between accuracy degradation and execution time improvement, we report the corresponding data for a comprehensive set of possible and meaningful regularization degrees. These results are graphically illustrated and allow an easy choice of the trade-off. It is generally a good idea to focus on Pareto optimal points with respect to accuracy degradation and execution time improvement in this data set for a first investigation.
Beyond the choice of the meaningful application of the degree of regularization, the level of maximal possible meaningful regularization can reveal the information of how well the data set, which is used for training is suited for this form of regularization. With the help of this, we determine a property, which we call \textit{regularization robust} on data sets, and identify properties, which make data sets more regularization robust.

Despite the realization of the regularization in scikit-learn \cite{pedregosa2011scikit}, we focus on a comprehensive experimental evaluation of the proposed regularizer in this paper. In detail, we take a set of UCI datasets \cite{UCI} and investigate the regularization in different model configurations. Furthermore, we conduct an extensive simulation study with synthetic datasets, where the relation between dataset properties and the effectiveness of the regularization is analyzed.
In short, we provide the following contributions:
\begin{itemize}
    \item A regularization term for the provocation of uneven splits in decision tree training, including an implementation in scikit-learn.
    \item Evaluation of the regularization on UCI datasets.
    \item An extensive simulation study with synthetic datasets to reveal the relation between dataset properties and regularization effectiveness.
\end{itemize}


\section{Related Work}
Performance optimization of trees and random forests is a widely studied topic in the literature. 
When it comes to concrete hardware-close implementations, one popular example is the \Cpp{} implementation for random forests in~\citet{wright2020ranger}.
The prominent concept of native trees, where nodes are stored in an array and executed in a narrow loop and if-else trees, where nodes form deeply nested if-else constructs, is introduced to maintain locality in the data and instruction memory in~\citet{Asadi/etal/2014}. 
More variances of tree implementations are studied for the runtime of inference on RISC-V MCUs~\cite{9925686}.\footnote{Its naming system is deviated to the terminology used by most of related works. For example, the if-else trees are named Naive kernels, and the native trees are named Loop kernels. In this work, we follow the majority and use the terminology commonly found in the literature.}

Random forests are also considered to be executed on accelerator devices, such as GPUs or FPGAs \cite{VanEssen/etal/2012,nakandala2020tensor, 7962153} or in a vectorized manner \cite{kim/etal/2010}.
In addition to the deployment of the models to a hardware-close language and massive parallel computation devices, also the optimization of the usage of the underlying hardware is investigated.
This includes optimization of the throughput in a pipeline execution~\cite{Prenger/etal/2013} and investigating the data structure and the decision tree structure itself and gain performance improvement with proper reordering~\cite{Dato/etal/2016,Lucchese2016}. 
More specifically, the usage of floating point hardware units and their performance impact is studied~\cite{hakert2022ecml,hakert2022flint}. 
\citet{chen2022efficient} utilize a probabilistic model of the data distribution in the data set to optimize the memory layout, in order to favor frequently used paths for the cache behavior. 

Although the approaches above provide various optimized implementations of random forests, they do not alter the training process in order to gain execution time performance.
One relevant approach is hyperparameter tuning \cite{Bischl2023}. 
Hyperparameter tuning specifically for random forests is covered in \citet{probst2019hyperparameters}, resulting in the tool \texttt{tuneRanger} focusing on both accuracy maximization and explainability. 
The tool, however, does not include execution time performance as an objective.
Mondrian forests \cite{lakshminarayanan2014mondrian} in contrast, introduce an online adaptive realization of random forests, which can improve the execution time performance while maintaining a similar accuracy.

Regularizing the training process of random forests, to the best of our knowledge, has not been studied for the objective of execution time performance. Regularization, also beyond the scope of random forests, however, is a studied topic in order to provide more explainability \cite{wu2018beyond} or achieve higher accuracy \cite{scheffer2000nonparametric}. 
Also, the effect of high randomness in the random forest training as a form of regularization is investigated \cite{mentch2020randomization}. 
The objective of error tolerance and robustness is further shown to be addressable by regularizing the training of binarized neural networks (BNNs) \cite{9473918}.

\section{Tree Regularization}

Improving the execution time of decision tree inference on real hardware opens a larger design space. One way to achieve faster inference is to decrease the size of the model itself. The obvious benefit towards execution time is, that less computation is needed to return an inference result. This approach usually introduces degradations in accuracy, as such the model cannot be shrank to an arbitrarily small size. A widely used method to decrease the model size is limiting the maximal depth a tree is allowed to grow to.

In this work, we introduce an alternative method towards reducing the overall model size. We optimize the decision tree construction to increase the existence of uneven splits to benefit shorter paths to leaf nodes.
More precisely, a penalty term in the splitting criterion is introduced, which serves as a control parameter to trade-off between tree size and predictive accuracy. This control parameter effectively shrinks the model size and reduces depths of single paths by maintaining and provoking uneven split decisions.

Another effective method is utilizing the cache behavior of the CPU. Chen et~al. have shown that reordering the nodes inside memory in a cache-friendly manner improves execution time \cite{chen2022efficient}.
In their approach the split probabilities of nodes are used to determine the new order in memory. Here uneven splits are beneficial, as they result in nodes which are accessed more often. Therefore, the benefit of the cache-friendly ordering is increased. Our proposed regularization both optimizes for smaller model sizes and increases the likelihood of uneven splits.

For the sake of completeness, we first give a short overview of the decision tree construction with the CART algorithm. Afterwards, we present the introduced regularization and how it can be tuned for different scenarios iteratively. Lastly, we discuss why the persistence of uneven splits are orthogonal to the cache-aware optimizations in detail. 


\subsection{Decision Tree Construction}

A widely used training method to construct decision trees is the CART algorithm~\cite{breiman1984cart}, by which the samples are repeatedly split by a chosen criterion. The result will be two sets of samples from the dataset. This is recursively repeated until a given stopping criterion is met (e.g. a certain depth is reached).

The basic working principle of all split criteria is to compute a score for all possible split values at each node, and then select the split point corresponding to the best combined criterion scores in the two resulting child nodes. 
More precisely, for a classification problem with $k$ labels, $p_i$ denotes the proportion of samples with class $c_i$ ($i = 1, \ldots, k$) in a node.
A widely used score for the impurity is the Gini impurity, which is measured as
\begin{equation}
	\operatorname{GINI}=1-\sum_{i = 1}^k p_i^2.
	\label{eq_gini_base}
\end{equation}
Hence, when all samples belong to one class, the sum is 1 and the resulting impurity is 0.
The Gini impurity results in a larger value, the more evenly the class labels are distributed in the node.
One popular way to find the best split inside a node is finding the minimal mean Gini impurity of both resulting child nodes. We note that other split criteria such as Entropy and Information Gain can also be used \cite{breiman1984cart}. However, as the exact criterion for splitting is not relevant for our proposed regularization, only the Gini impurity is covered here.


The split results in the samples being separated into two portions, being further used in the left and right child. This division in the samples then determines the probability of the left or right subtree to be used in an inference, 
Each node has a distinct access path starting from the root node and ending in the node itself. To get the absolute probability of any node, the individual probabilities of every node on the path to that node need to be multiplied. The resulting value is the probability of this node to be accessed during prediction. Intuitively, the absolute probability of the root node is 100\%. 
The probability of any path to be taken during inference is the probability of the leaf node the path ends in. These absolute probabilities can be used to identify which paths are frequently accessed.

\subsection{Regularization Factor}
\label{subsec:regularizer}

A possibility to improve the execution time is to reduce the total model size by controlling the training process to only keep important paths. The reduced amount of nodes leads to less memory loads during an inference. In addition, this regularization of the training can be designed such that not only important paths in terms of prediction accuracy are kept, but also the access frequency of paths is maintained kept.
This consequently leads to an orthogonal optimization of the cache optimization from Chen et~al., since the cache friendly handling of frequently accessed paths is kept, and cache replacements are reduced.

Since training of decision trees according to the CART algorithm~\cite{breiman1984cart} consists of recursively splitting the samples into two child nodes based on a threshold value, the split decision can be modified in order to favor asymmetric probabilities.
In order to allow a trade-off between the original split criterion and the size-aware split, we introduce an additive regularization factor for the split criterion, penalizing even splits.
The amount of penalization can be controlled with a real-valued factor $\lambda$ which is subject to tuning. Although this design is applicable to arbitrary split criteria, we here restrict ourselves to the popular Gini impurity criterion in order to analyse the effect in depth.

In order to include a size-aware splitting criterion into this process, we define a regularization term $R$ as
\begin{equation}
	\label{eq_regularizer}
	R=1-\frac{|\#\text{samples}_{\text{left}}-\#\text{samples}_{\text{right}}|}{\#\text{samples}}.
\end{equation}
Hence, when the split distributes samples almost equally to the left and right child nodes, the value is close to 1, when the split is very asymmetric on the other hand, the value is closer to 0.
Note that in contrast to the Gini impurity, the regularization term does not operate on the class labels, but instead on the number of samples.
In order to form the resulting split criterion, we add the regularization term with an adjustable weight $\lambda \in \mathbb{R}^+$ to the Gini impurity:
\begin{equation}
\operatorname{GINI}'=\operatorname{GINI} + \lambda \cdot R.
\label{eq_gini_mod}
\end{equation}
Adding the regularization term to the evaluation and optimization of the Gini impurity in every step of the CART algorithm allows accounting for cache-friendly splits during the training. 
It should be noted that the introduction of the regularization potentially degrades the Gini impurity and hence also the accuracy of the trained model.
Consequently, the parameter $\lambda$ has to be chosen effectively to provide a good trade-off between accuracy and asymmetric splits.

Our modifications are directly implemented in scikit-learn. 
To achieve the outlined regularization, a new split criterion based on the Gini impurity is introduced. The implementation is largely similar as for the standard Gini split criterion. However, when calculating the node impurity, the resulting value is adapted according to \Cref{eq_gini_mod} and returned. To accommodate the factor $\lambda$, an additional hyperparameter can be set while fitting the model to control the amount of regularization. The source code is publicly available under \url{[hidden due to double blind submission]}.
\subsection{$\lambda$  Tuning}
\label{sec:tune}

During training, the regularization factor $\lambda$ needs to be set. It should improve training towards the best performance optimization while preserving the accuracy as good as possible. An optimal regularization factor cannot be picked universally. The effectiveness and influence of the factor changes highly depending on a variety of factors (e.g., the number of classes in the dataset).

There is a limit to how much any split can be usefully regularized, as at some point all samples would go to one child node.
Therefore, the impact of the regularization factor is going to approach a limit the larger the factor gets.

To find the optimal factor for a given scenario, the expected performance improvement needs to be quantified.
To that end, we define the expected depth of a single tree. It is measured as
 \begin{equation}
    \sum_{l \in \text{leaf}(t)} p_l * \text{depth}(l)
\end{equation}
where $\text{leaf}(t)$ are all leaves of tree $t$, $p_l$ is the probability of leaf node $l$ and $\text{depth}(l)$ is the depth of node $l$. 
The expected depth is therefore the mean depth the inference is expected to reach during repeated inference operations. Consequently, a reduction in the expected depth results in an increase in performance, as fewer nodes have to be loaded during inference. Furthermore, once the expected depth does not change significantly, the influence of the regularization factor is less pronounced and less performance gain is to be expected. To find an optimal factor, the factor is iteratively increased until the difference in expected depth falls under a set threshold, which decides how close to the best possible performance improvement the factor is tuned.
At that point, performance is unlikely to improve further, and the corresponding value for $\lambda$ is chosen.


\section{Experimental Evaluation}
To evaluate the application of the hardware-aware regularization, we conducted experiments on real and synthetic data sets. First, we apply a default setting, where the maximal regularization is applied with a configurable, tolerable accuracy degradation. Second, we enlighten the trade-off between degree of regularization, speed improvement and accuracy drop. Lastly, we evaluate the limitations of regularization itself and report the boundaries for the meaningful application.



\subsection{Evaluation Setup}
\label{sec:setup}
For evaluating the execution time improvement, we trained random forests with different degrees of regularization (i.e. varying $\lambda$) on real and synthetic datasets. We subsequently generated a straightforward C implementation and a cache optimized implementation via \citet{chen2022efficient}. The generated trees of both implementations are executed on a real world target machine. We use a server class system, i.e. with an Intel(R) Xeon(R) Gold 5218 @ 2.3GHz CPU with 16 cores, 1024 KiB L1 Cache, 16 MiB L2 Cache and 22 MiB L3 Cache and 180GB RAM. We utilized Scikit-learn to train random forests with varying number of trees and maximal tree depths (1, 5, 15, 20) for each of these datasets. After a given threshold, the number of trees was only increased for more experiments if it improves the accuracy enough. This was done to reduce the amount of redundant experiments. 
To provide a better intuition for the impact of the regularization, we always compare the regularized implementation to the comparable not regularized counterpart. In greater detail, the not cache optimized regularized implementation for a specific number of trees and maximal tree depth is compared to the not regularized version of the not cache optimized implementation. This is similarly done for the cache optimized implementations.

 In addition, we measured the balanced accuracy of the trained model based on the test dataset. Balanced accuracy evaluates a model's classification performance by considering both sensitivity (true positive rate) and specificity (true negative rate), making it particularly useful in scenarios with imbalanced datasets. Since the methods from \citet{chen2022efficient} only optimize the memory layout and do not change the model structure, the balanced accuracy is the same for all implementations.  For the measurement of the execution time, we executed 50 repetitions of the inference of the test dataset and average the time consumption under realistic execution conditions.
To compare the balanced accuracy and mean relative execution time, we used a training-test split ratio of 3:1 and repeated it 8 times. The scitkit-learn hyperparameter $max\_features$ was varied across a set of recommended default values $\{\frac{\lfloor\sqrt{p}\rfloor}{2}, \lfloor\sqrt{p}\rfloor, 2\lfloor\sqrt{p}\rfloor, p \}$, where $p$ denotes the number of features \cite{wright2020ranger, hastie2009elements, liaw2002classification}.

\subsection{UCI Datasets}
\label{sec:data}

In the following experiments, the influence of the regularization is compared on eleven datasets from the UCI repository, which was also adopted in \citet{chen2022efficient}. 
Table \ref{tab:datasets} lists the dataset name, source, number of samples ($n$), number of features ($p$), and number of classes ($cl$). 
\begin{table}[h!]
    \centering
     \caption{Name, source, number of samples ($n$), number of features ($p$), number of classes ($cl$) of each used dataset.}
     \scriptsize
       \begin{tabular}{llccc}\hline
         Dataset&Source&$n$&$p$&$cl$ \\\hline
         Adult&\cite{adults}&48,842&64&2\\
         Bank Marketing&\cite{bank}&45,211&59&2\\
         Covertype&\cite{covertype}&581,012&54&7\\
         Letter&\cite{letter}&20,000&16&26\\
         Magic&\cite{UCI}&19,020&10&2\\
         MNIST&\cite{UCI}&45,000&784&10\\
         Satlog&\cite{UCI}&6,435&36&6\\
         Spambase&\cite{UCI}&4,601&57&2\\
         Sensorless Drive&\cite{UCI}&58,509&48&11\\
         Wearable Computing&\cite{wearable}&165,632&17&5\\
         Wine Quality&\cite{wine}&6,497&11&7\\\hline
    \end{tabular}
 
    \label{tab:datasets}
\end{table}
For the ease of presentation, we aggregated the multiple simulation settings with regard to the 8 replications and 
focused on results obtained with $max\_features$ set to $\lfloor \sqrt{p} \rfloor$. 
Detailed results for all settings are available in the Appendix.
\subsubsection*{Intuitive Application}

\begin{figure*}[ht]
    \centering
    \includegraphics[width=1\textwidth]{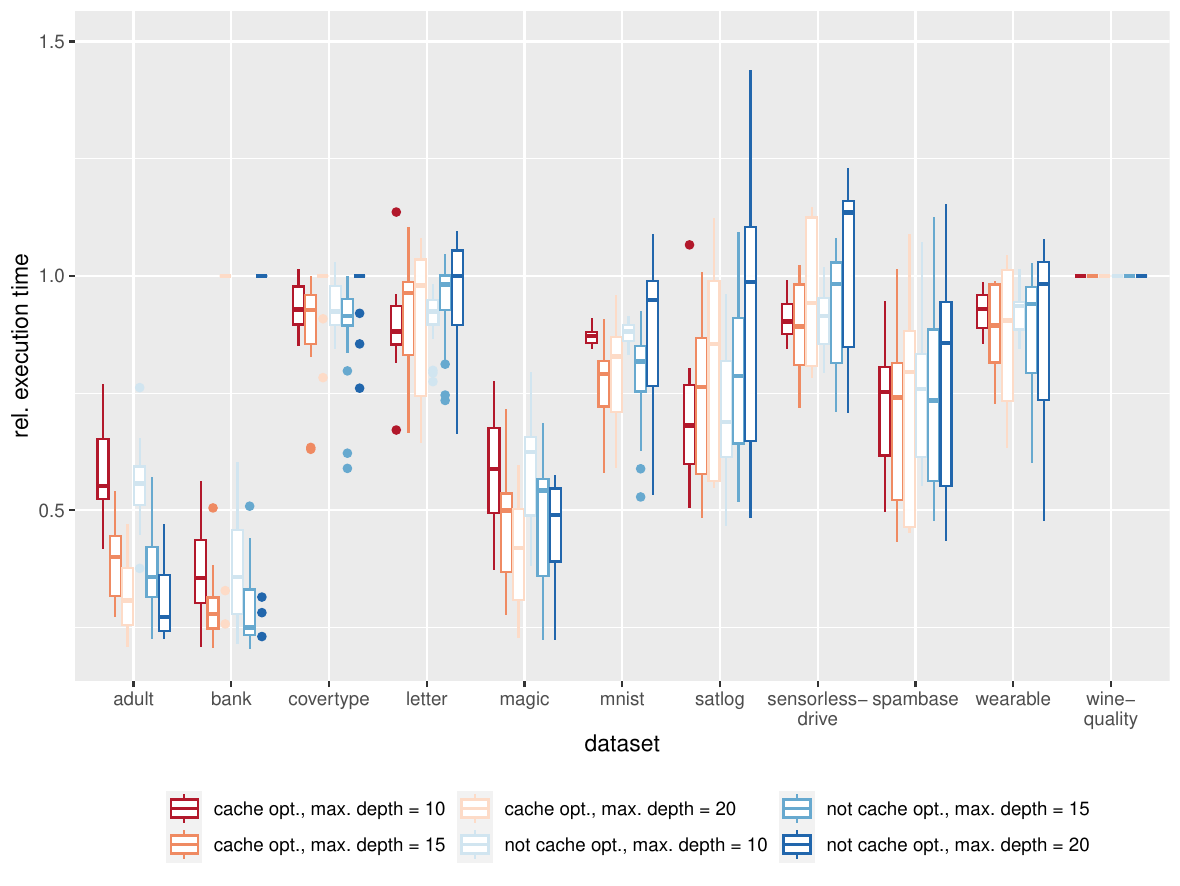}
    \vspace{-1cm}
    \caption{Impact of regularization on execution time across datasets}
    \label{fig:pic1}
\end{figure*}

To illustrate the most intuitive use case of the regularization, we limit the allowed degradations in accuracy to 5\%. We then pick the best regularization factor $\lambda$, which achieves the maximal execution time improvement, while not degrading the accuracy beyond the specified level.
\Cref{fig:pic1} reports the corresponding results, where the x-axis separates the different data sets from the UCI repository. The y-axis shows the relative speed improvement with regularization in comparison to the same configuration without regularization. Each box includes random forests with different numbers of trees.
The different colors indicate different maximal depths of the trained decision trees and configurations without and with cache optimization.

From the presented results in Figure~\ref{fig:pic1}, several observations can be made. First, it can be seen that for trees with a small maximal depth, the improvement in terms of execution time is not reliably observable. Some configurations degrade the speed, some configurations only slightly increase the speed. Considering that a limited maximal depth of $1$ only allows for 3 tree nodes, these results are not surprising. Further, it can be observed that the speed improvement grows, the deeper the trees become. A general tendency can be observed, that the deepest trees also benefit most from regularization in terms of execution time improvement. 
For the data sets, which achieve a significant execution time improvement, a similar scale of improvement for not cache optimized and cache optimized implementations can be observed. This supports the design principle of a regularization, improving both not cache-optimized and cache-optimized implementations in an orthogonal manner. It should be noted that this plot shows the relative execution time in comparison to the not regularized version, i.e., when the cache optimization improves the execution time upon the not cache optimized implementation, this improvement is orthogonal to the regularization. 
The maximal improvement in terms of execution time can be observed to be more than $75\%$, i.e., more than $4\times$ faster than without regularization. The data sets, which profit most from the regularization in terms of execution time improvement are adult, bank and magic. Spambase and satlog also show a higher timing improvement than most of the other data sets. Comparing this finding to \Cref{tab:datasets} suggests the conclusion that data sets with binary classification can benefit most from the regularization in terms of execution time improvement.

\subsubsection*{Regularization Trade-Off}
\begin{figure*}[ht!]
    \centering
    \includegraphics[width=1\textwidth]{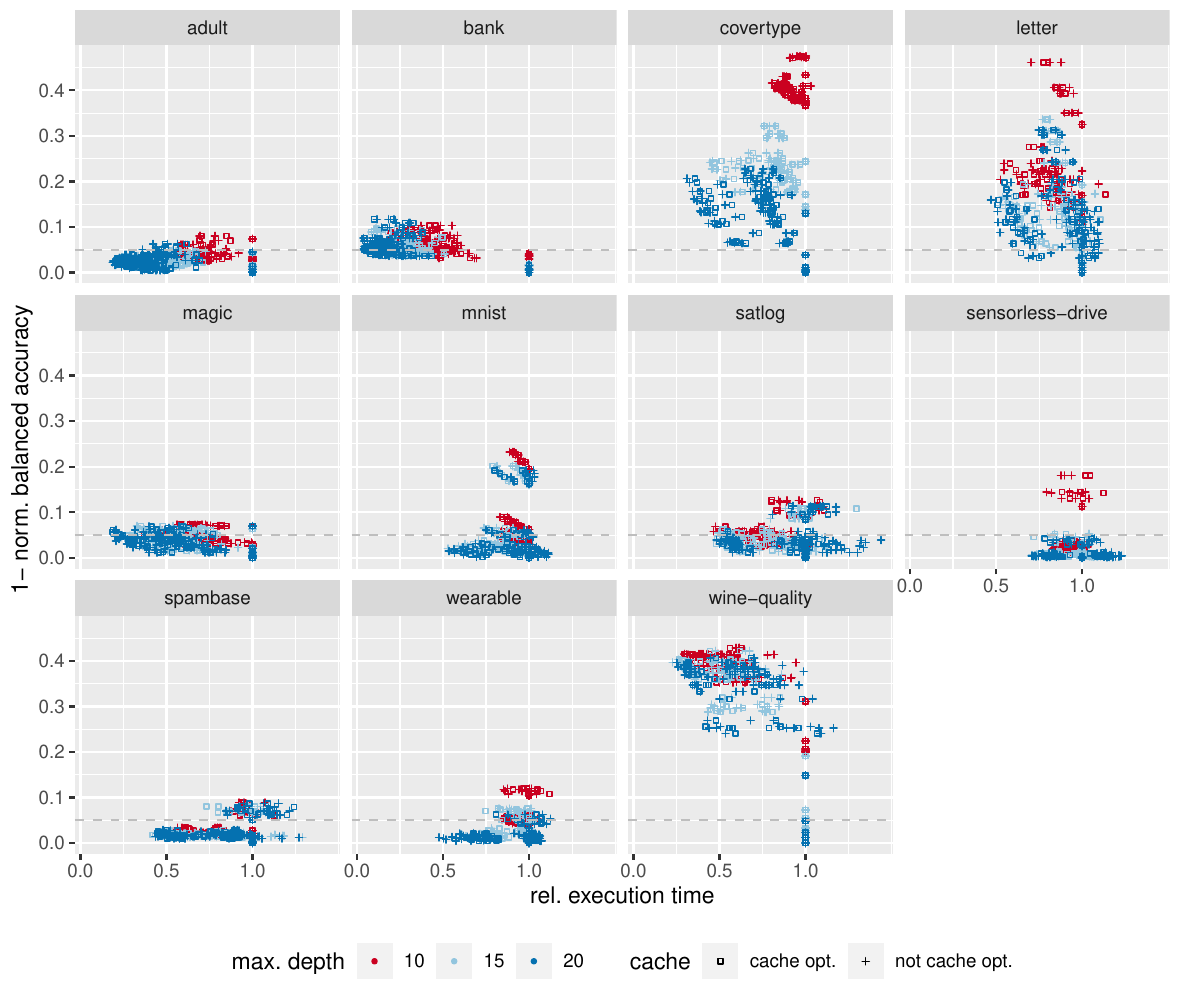}
    \vspace{-1cm}
    \caption{ Normalized balanced accuracy values (with respect to the maximum value within each data set) and relative execution times for all data sets with different tree depths and separated by cache optimization or not.}
    \label{fig:pareto}
\end{figure*}

Tolerating only accuracy degradation until a configurable threshold is a simplified form of application, which does not allow to make a trade-off.
It could still happen, that a higher degree of regularization degrades the accuracy slightly beyond this threshold, but achieves significant faster speed. Such scenarios are evaluated by analyzing the relation between the accuracy drop and the runtime improvement for different degrees of regularization.
We illustrate the results of corresponding experiments in \Cref{fig:pareto}. The data sets are separated in different subplots. Each configuration, including different amounts of trees and different degrees of regularization.
forms one point, which is denoted by the relative execution time improvement to the corresponding not regularized counterpart on the x-axis and the accuracy drop on the y-axis respectively. Cache optimized and not cache optimized implementations are separated by squares and pluses. We further denote the limit of $5\%$ accuracy drop, as used for the previous intuition, by a dotted gray line.

From the results, two different major behaviors can be identified: For certain data sets, namely adult, bank, magic, mnist, spambase and wearable, the configurations with the maximal speed improvement are either Pareto optimal or only have a slight larger degradation in the accuracy than the configurations with the lowest accuracy degradation. This trend can be observed to exist across different maximal depths of trees. For the other configurations, it can be observed that a higher execution time improvement also comes with higher accuracy degradation, especially for deeper trees.
It can be as well observed, that cache optimized and not cache optimized implementations form close results, which supports again the design principle of an orthogonal optimization.
This suggests the conclusion that, data sets with either binary classification or large sample sizes are better suited for execution time improvement due to regularization without high accuracy impact than other data sets. We call these data sets \textit{regularization robust}.

\subsubsection*{Limits of Regularization}
\begin{figure*}[ht]
    \centering
    \includegraphics[width=1\textwidth]{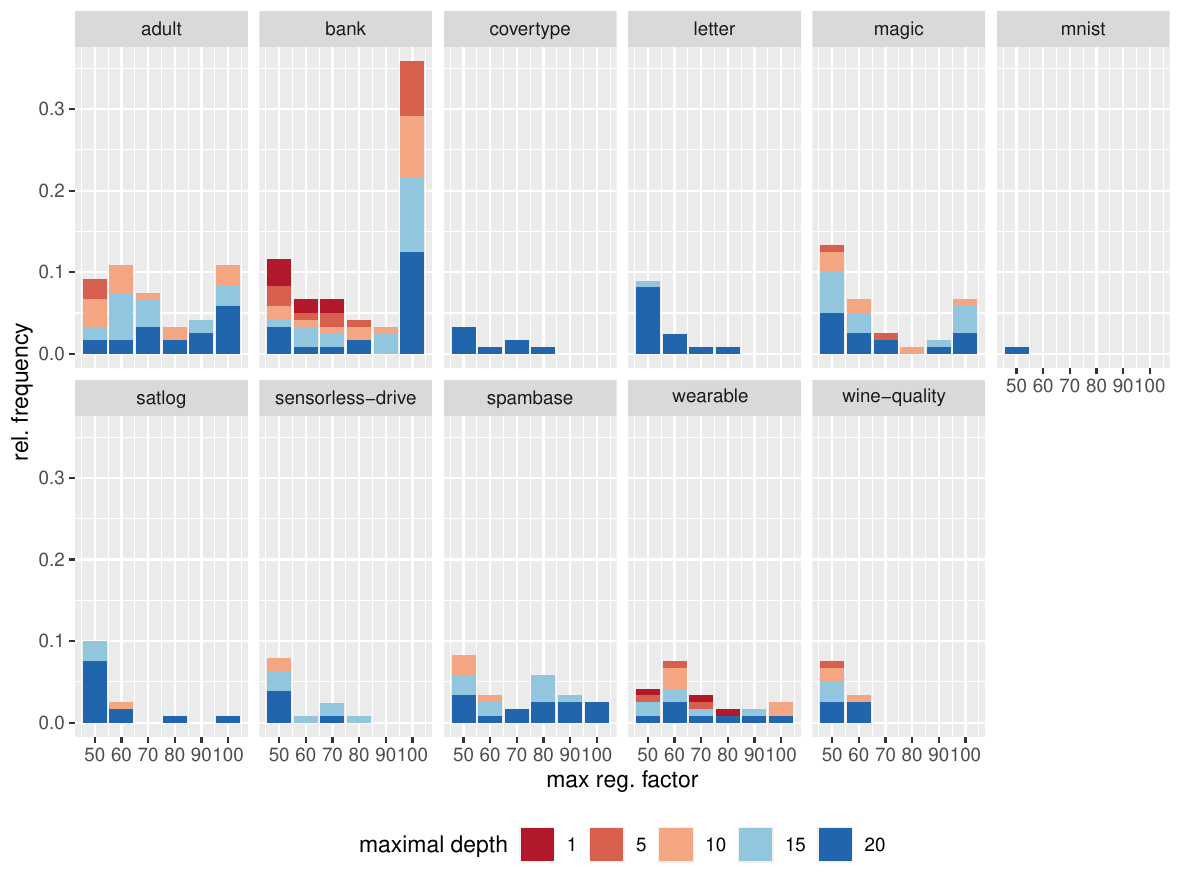}
    \vspace{-1cm}
    \caption{Relative frequency of maximum regularization factors in each data set, focusing on the range of values above 40.}
    \label{fig:MaxFreq}
\end{figure*}
The previous experiments and result discussions suggest the conclusion, that several data sets have a property, namely \textit{regularization robust}. 
This property refers to data sets, which can benefit strongly from high degrees of regularization in terms of their improved runtime, while not sacrificing too much accuracy. Previous experiments suggest that the data sets adult, bank, magic, mnist, spambase and wearable have this property to a certain degree. In order to investigate this property even further, we use regularization tuning as explained in \Cref{sec:tune} to stop increasing the regularization factor once the expected increase in speed falls under a change threshold of 5\%.
In other words, if no further execution time improvement is achieved, the regularization factor is not further increased. 

We illustrate the amount of configurations of a data set (i.e. different number of trees in an ensemble and different implementation strategies) with their maximal regularization factor in \Cref{fig:MaxFreq}.
To get a general picture of the influence of the regularization factor $\lambda$, experiments with $\lambda \in [0, 40]$ are run regardless of any metric. Next, regularization tuning is used to stop once the factor is expected to not make a significant difference to execution speed.
It can be observed that for certain data sets, a reasonable amount of configuration can profit from high regularization factors. These data sets are adult, bank, magic, spambase and wearable. Except the mnist data set, which has an exceptional high number of classes, this is exactly the list of data sets, which are encountered as regularization robust before.
Hence, by only investigating the data set properties upfront, an assertion can be made whether the data set is regularization robust and thus may profit from strong degrees of regularization. We have seen that this often holds 
for binary classification problems or very large data sets.

\subsection{Simulation with Synthetic Data}
\label{sec:simu}

%
In order to better understand the effects of tree regularization on binary classification datasets, particularly with respect to \textit{regularization robustness}, we conducted experiments with synthetic data. We first describe how the data is generated and subsequently present the measured results.

\subsubsection*{Simulation Setup}

%
%

We consider a binary classification problem $Y\in \{0, 1\}$ with ten real-valued features $X_1,\ldots, X_{10}$ for which we specify different distributions, dependencies and underlying models, described in the following.
We model the last five features $X_6,\ldots ,X_{10}$ as independent and uniformly distributed random variables from ${[0,10]}$, independent of the first five features $X_1,\ldots, X_5$. For the first five features we consider three different dependence structures as summarized in Table~\ref{tab:festures}.  In the first setting (Independent), we consider completely independent features, where each feature follows a distinct mixed normal distribution $Z_{\ell,k}= l W_1 + (1-\ell) \mathcal{N}(\Delta_\mu +k, 1)$ with  $\Delta_\mu \in \{1, 3, 5, 8\}$ and  $W_1\sim \mathcal{N}(1,1)$. The concrete choices for $\ell$ and $k$ are given in Table~\ref{tab:festures}. In the other two settings, we model a weak dependence between $X_2, X_3$ and $X_4$ (second last column) and a strong dependence between $X_1$ and $X_5$ (last column), respectively. Note that for the first feature, $k = b$ is not held constant, but is systematically varied from 0.2 to 0.9 to provide some degree of adjustment to modulate the class balance.
\begin{table}[h!]
    \centering 
    \caption{Distributions of $X_1,\ldots, X_5$ used in the simulation studies with \mbox{$p\in \{0.2, 0.5, 0.7, 0.9\}$}.} 
    \label{tab:festures}
    \scriptsize
   \begin{tabular}{cccc}
\hline
Features&Independent&Weakly dependent & Strongly dependent\\\hline 
$X_1$&$Z_{p, 1}$&$Z_{p, 1}$&$Z_{p, 1}$\\
$X_2$&$Z_{0.1, -5}$&$Z_{0.1, -5}$&$Z_{0.1, -5}$\\
$X_3$&$Z_{0.5, 2}$&$Z_{0.5, 2}$&$Z_{0.5, 2}$\\
$X_4$&$Z_{0.3, 3}$&$Z_{0.1,-5}+Z_{0.5,2}$&$Z_{0.1,-5}+Z_{0.5,2}$\\
$X_5$&$Z_{0.8,-2}$&$Z_{0.8,-2}$&$Z_{0.5,2}+0.5Z_{p,1}$\\
\hline\\
\end{tabular}
\end{table}
Having fixed the dependencies among the features, we now model the dependencies with the outcome $Y$. In this study, we investigate three different relationships between $Y$ and $X_1,\dots,X_{10}$ by incorporating different dependencies and correlation structures through logical rules. The settings range from a simple dependence of the outcome solely on the first feature $X_1$ (S1), whereby $Y$ equals 1 if the realization derived from $X_1$ originates from $W_1$ of the normal mixed distribution. The more complex dependencies involve the first three or five features. The concrete details are illustrated in Table~\ref{tab:models}.
\begin{table}[h!]
    \centering 
    \caption{Dependent models between the output and some of the features. Here $O_i$ refers to the event that the realization of the feature $X_i$ originates from the first part of its mixed normal distribution.} 
    \label{tab:models}   
   \begin{tabular}{ll}
   \hline
   Setting& $Y=1$\\\hline
S1  &  $O_1$ \\ 
S3& ($O_1$ and $O_4$) or $\neg O_2$ \\
S5& ($O_1$ and $\neg O_3$) or ($\neg O_5$ and $\neg O_4$) or $O_2$ \\\hline
\end{tabular}
 \end{table}

For each setting, we generated samples of size $num$ from the respective model with $num \in \{100, 200, 500\}$. 
The regularization strength $\lambda$ and the number of trees are varied as described in \Cref{sec:setup}. The same applies to the hyperparameter $max\_features$.
%
\subsubsection*{Results}
For ease of presentation, we focus on the most important results and general trends. Studying the simulation study results for all configurations, we observed that changes in the dependency structure of the feature, the relationship between features and outcome and the size of the inner bootstrap sample ($max\_features$) of the random forest had no large effect on the behavior of trees under the regularization. In comparison, the balance of the prediction classes, regulated by the balance parameter $b$ and $\Delta_\mu$, and the sample size $n$ were the driving forces for changes in the influence of the regularization.

Examining the effects of sample size, we find results consistent with those of the previous section. As the sample size increases, a greater improvement in execution time is observed along with a decrease in accuracy. Details can be found in the appendix

\begin{figure*}
    \centering
    \includegraphics[width=1\textwidth]{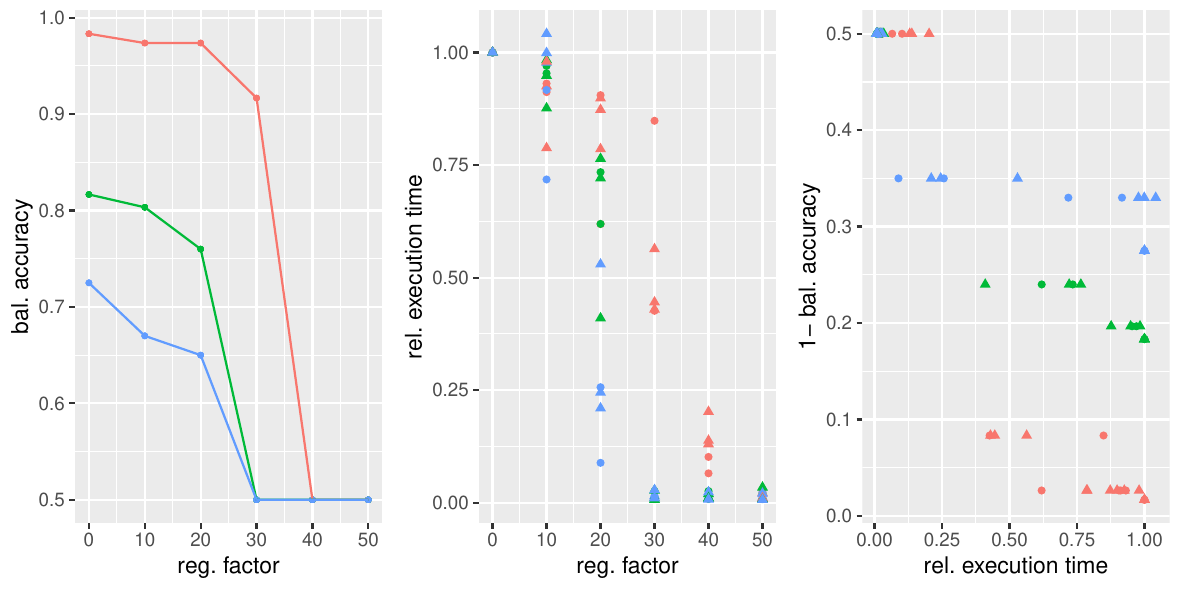}
    \vspace{-1cm}
    \caption{Simulation results of the regularization for varying $b$ and $\Delta_\mu$: $b=0.9, \Delta_\mu=8$ (red), $b=0.7, \Delta_\mu=3$ (green) and $b=0.5, \Delta_\mu=1$ (blue). The different shapes of the points indicate whether a cache-optimized version is used (circle) or not (triangle).}
    \label{fig:Simu}
\end{figure*}

Results for different combinations of the balance parameters are shown in Figure~\ref{fig:Simu}. The results shown in Figure~\ref{fig:Simu} are for $max\_features = 6, n = 100, num = 100,$
independent characteristics and the S3 model for the outcome. We present results for three combinations of $b$ and $\Delta_\mu$: $b=0.9, \Delta_\mu=8$ (red), $b=0.7, \Delta_\mu=3$ (green) and $b=0.5, \Delta_\mu=1$ (blue). 
These combinations were selected because of their different strengths of balance. Red is the most unbalanced and blue is the most balanced.
The x-axis for each of the first two plots shows an increasing regularization factor, the y-axis shows the balanced accuracy in \Cref{fig:Simu} (left), and the relative execution time in \Cref{fig:Simu} (center). The x-axis for the figure on the right shows the 1-balanced accuracy, and the y-axis shows the relative execution time. The different shapes indicates wether a cache-optimized version is used or not. For all three settings, there is a clear trend towards faster relative execution times as the regularization factor increases. However, the improvement diminishes as soon as the regularization factor reachs 50. When examining the effect of regularization on balanced accuracy in these settings (plot on the left), it is noticeable that the blue and green settings show a more significant decrease in accuracy than the red settings. This suggests that the balance of classes influences the effect of regularization, with unbalanced classes showing greater sensitivity to regularization. By looking at the Pareto front (plot on the right), we can see that the red setting dominates the others for most configurations.
\subsection{Discussion}
\label{sec:discussion}

The previously presented results indicate that the introduction of regularization offers a trade-off between a degradation of accuracy and
the improvement of execution time. While for  shallow decision trees the regularization generally cannot offer a large spectrum for the trade-off and quickly degrades to extreme cases, a wider spectrum for the trade-off is offered for deeper tree models in general. It is worth noticing that the degradation of the accuracy is usually less by one order of magnitude than the gained speed improvement, when a moderate amount of regularization is chosen.

Investigating the dataset properties itself, the comparison between synthetic and real data sets shows that a major influence on the effectiveness of the regularization is put by the balance of the prediction classes.
\Cref{tab:chis} shows the Chi-Squared values for the UCI datasets, where a high values indicate a potential high imbalance of the distribution in the prediction classes. 
\begin{table}[htb]
    \centering 
    \caption{Chi-Square Values of UCI Datasets}
    \label{tab:chis}
\begin{tabular}{l l}
\hline
 Dataset    & Chi-square  \\\hline
       \textbf{Adult}&\textbf{6,556.066}\\ 
       \textbf{Bank Marketing}&\textbf{18,552.56}\\
         Covertype&71,3450.9\\
         Letter&19.3273\\
         \textbf{Magic}&\textbf{1271.086}\\
         MNIST&139.705\\
         Satlog&504.379\\
         \textbf{Spambase}&\textbf{153.515}\\
         Sensorless Drive&2.2707\\
         Wearable Computing&33,934.62\\
         Wine Quality&6,383.254\\\hline     
\end{tabular}
\end{table}
It should be noted that these values can only be interpreted for binary classification datasets (indicated in bold), since the structure of the imbalance becomes too complex for multi label classifications. It can be seen, that for the adult and bank dataset, which are observed to provide high speed improvements on minimal accuracy degradation, a relatively high Chi-Square value can be observed. This aligns with the observations from the synthetic datasets, where highly imbalanced distributions also allow high speed improvements on minimal accuracy degradations.


From the perspective of an user, regularization should be considered for deeper tree models, since the effectiveness for small trees is highly limited. When a dataset is used, which by default is imbalanced, regularization can be generally turned up further and gain more speed improvement while not degrading the accuracy much.


\section{Conclusion and Outlook}
Deploying machine learning models efficiently on resource-constrained devices requires a carefully tuned model shape in terms of model size and a hardware-close implementations, of which the state-of-the-art cache-aware optimizations are prominent for random forests and decision trees. 
In this work, we present a method to regularize the impurity computation and reward highly asymmetric distributions in the training process of decision trees, which provokes uneven probability distributions (i.e., uneven splits) for offering high optimization potential.

To examine the effectiveness of our method, we conduct extensive experimental evaluation on synthetic datasets and on UCI datasets. 
The evaluation results show that a large execution time reduction of up to $\approx 4\times$ can be gained in many cases while degrading the target accuracy by a few percent. The user can either specify an acceptable threshold of sacrificable accuracy degradation and derive the optimal regularized result or can make an own trade-off by choosing between pareto optimal points in the scope of accuracy degradation and execution time improvement.
We can further categorize data sets as regularization robust, when they are either binary classification data sets or have a high amount of samples. Such data sets may benefit strongly from regularization.
Spending a deeper focus on the property of being regularization robust, we see a dependency to the sample size in the synthetically generated data sets. We further observe a strong dependency with the imbalance of the synthesized data sets and the effectiveness of regularization, supporting the initial design principle.
An implementation in scitkit-learn is openly available.

For future work, the application of regularization across the random forest structure should be studied, instead of considering single trees in separation. For instance, the dataset can be split into subsets with strong dependencies for the training of different trees, making the regularization more effective. Furthermore, it can be considered to have heterogeneous degrees of regularization for other tree-based ensembles.

\bibliography{main}
\bibliographystyle{mlsys2024}

\clearpage
\section*{Appendix}

We present the complete results regarding $max\_features$ and the different execution types. The results are presented in graphical form.

\begin{figure*}[ht]
    \centering
 \includegraphics[width=0.7\textwidth]{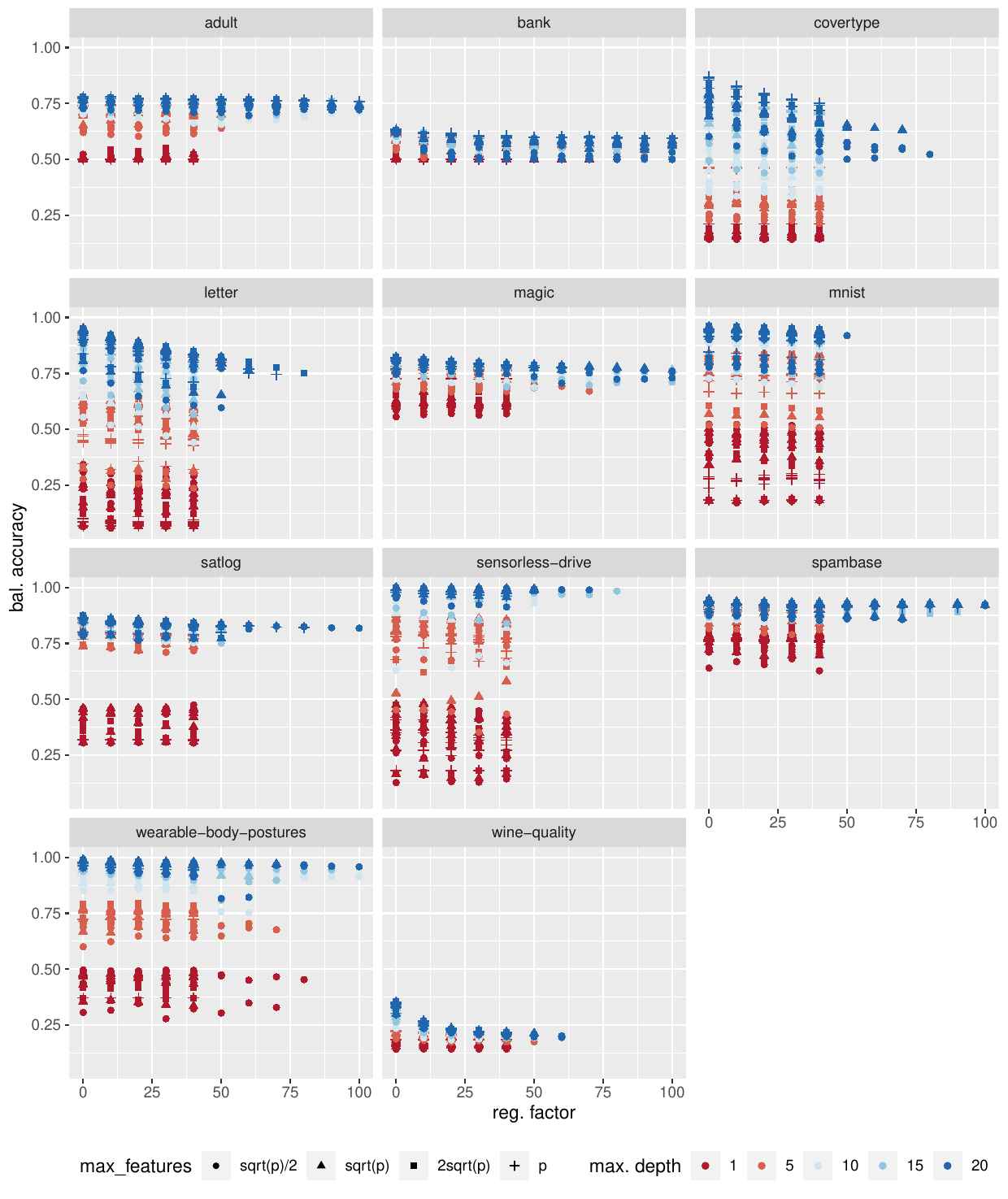} 
 \caption{Evaluation of the balanced accuracy for the UCI datasets}
\end{figure*}

\begin{figure*}[ht]
    \centering
 \includegraphics[width=1\textwidth]{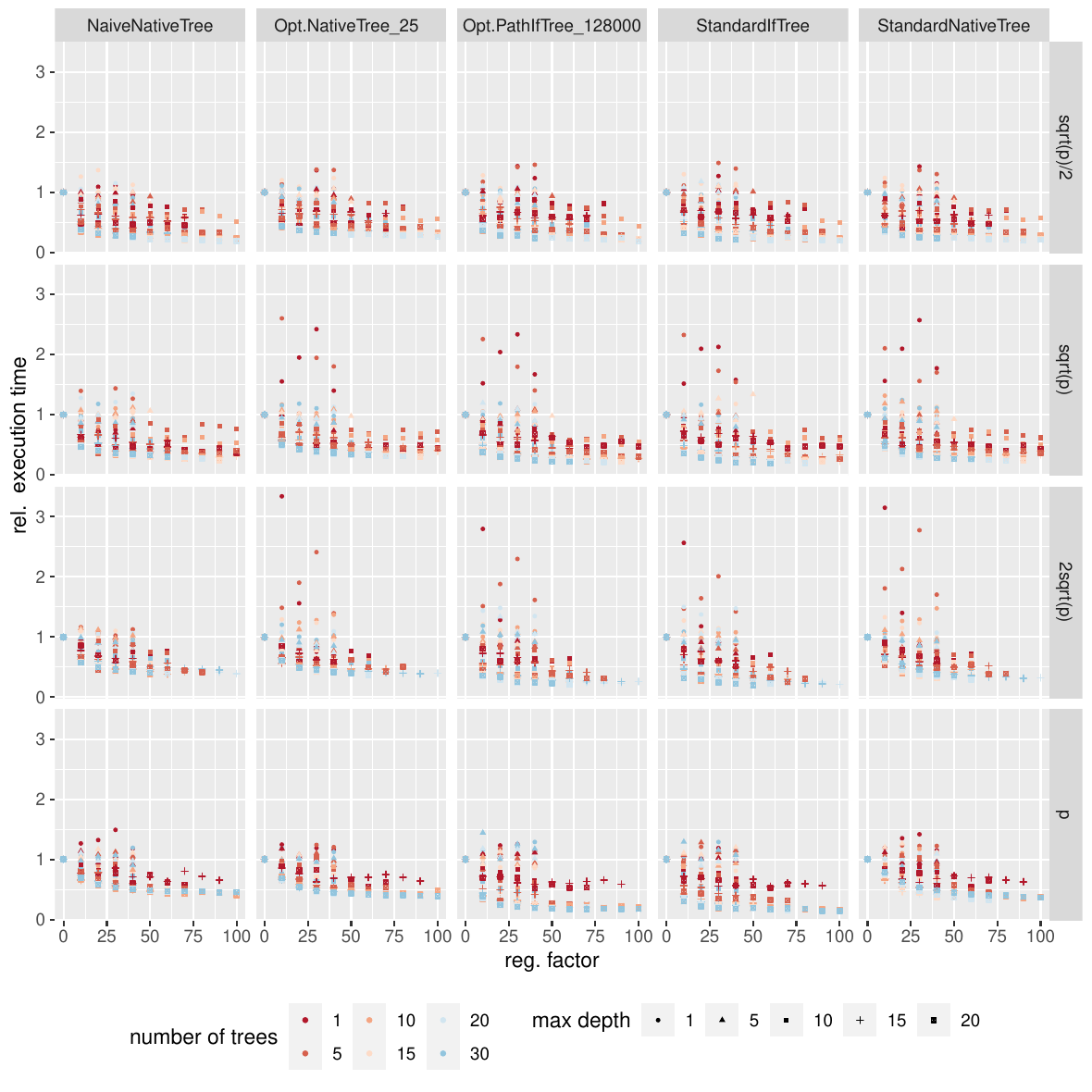} 
 \caption{Evaluation of the relative execution time for the adult dataset separated by the execution type $max\_features$, the maximum depth (shape of the points) and the number of trees (color).}
\end{figure*}

\begin{figure*}[ht]
    \centering
 \includegraphics[width=1\textwidth]{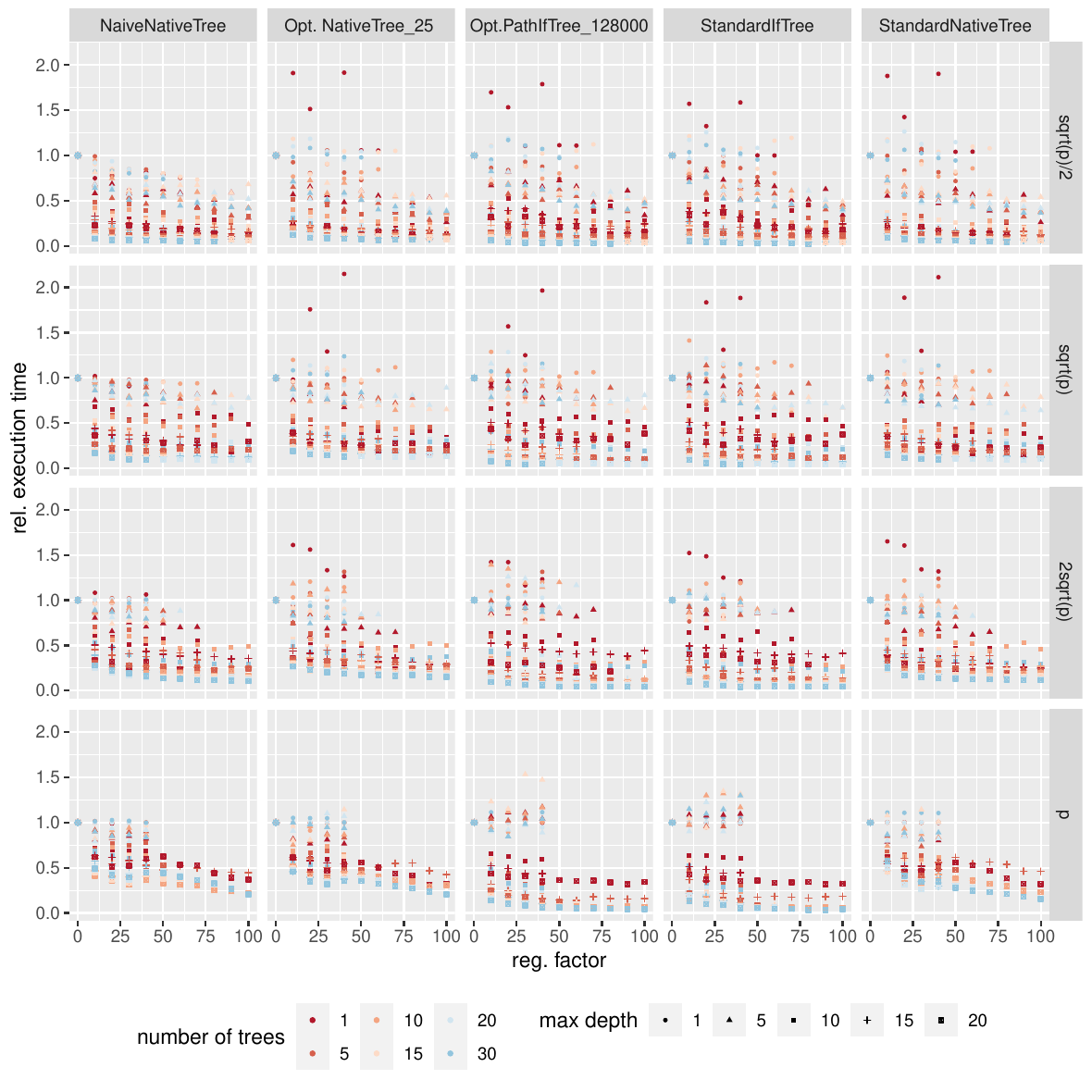} 
 \caption{Evaluation of the relative execution time for the bank dataset separated by the execution type $max\_features$, the maximum depth (shape of the points) and the number of trees (color).}
\end{figure*}
\begin{figure*}[ht]
    \centering
 \includegraphics[width=1\textwidth]{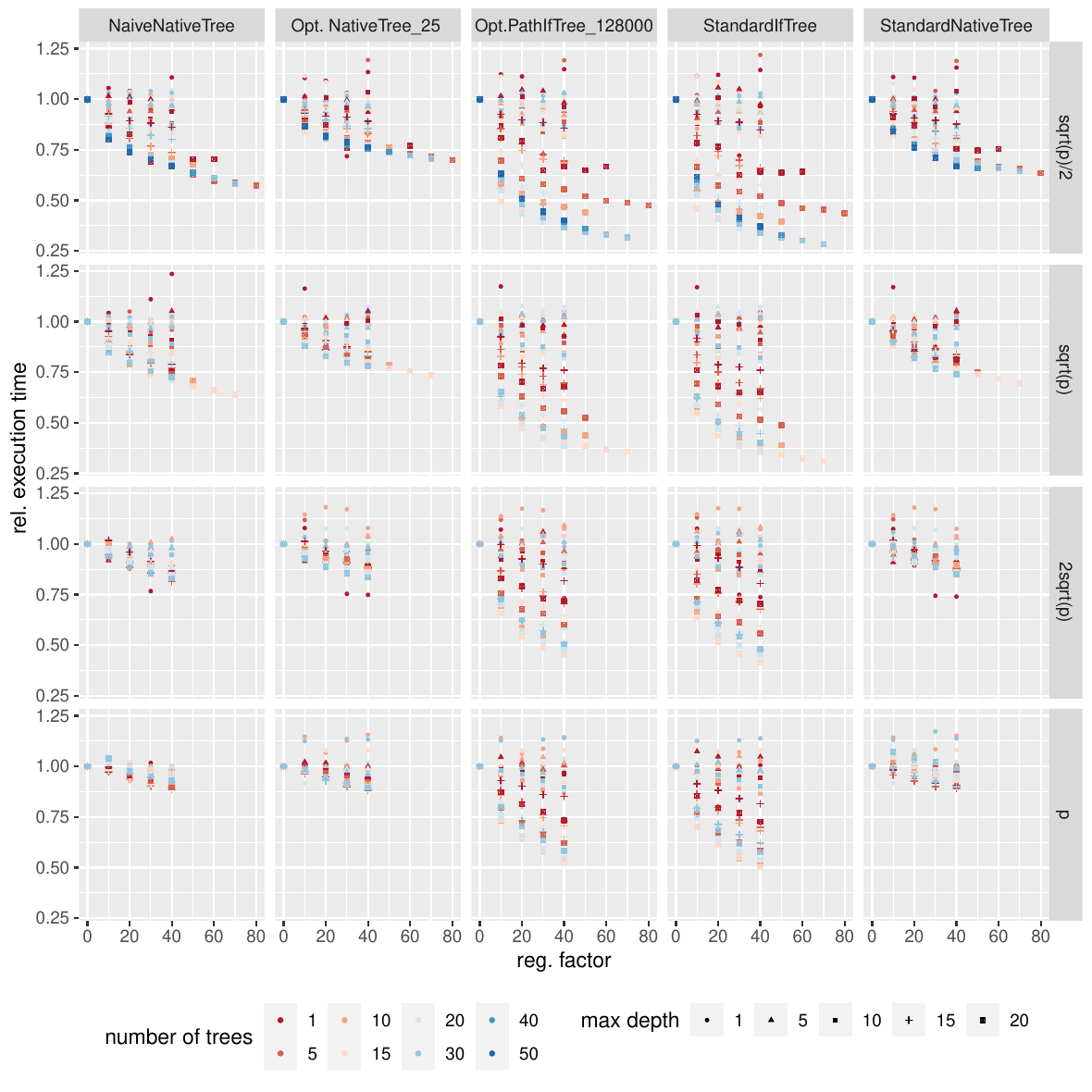} 
 \caption{Evaluation of the relative execution time for the covertype dataset separated by the execution type $max\_features$, the maximum depth (shape of the points) and the number of trees (color).}
\end{figure*}
\begin{figure*}[ht]
    \centering
 \includegraphics[width=1\textwidth]{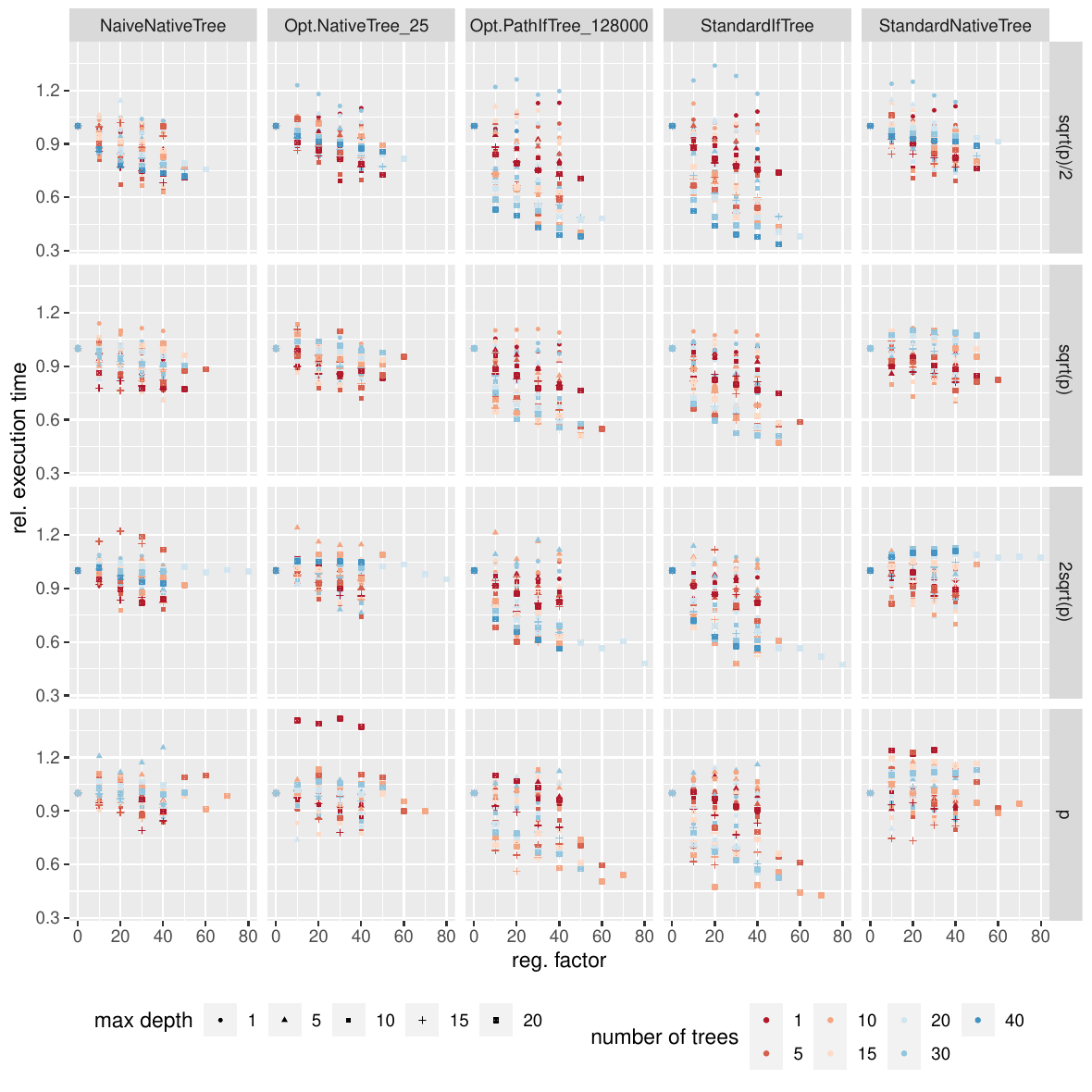} 
 \caption{Evaluation of the relative execution time for the letter dataset separated by the execution type $max\_features$, the maximum depth (shape of the points) and the number of trees (color).}
\end{figure*}
\begin{figure*}[ht]
    \centering
 \includegraphics[width=1\textwidth]{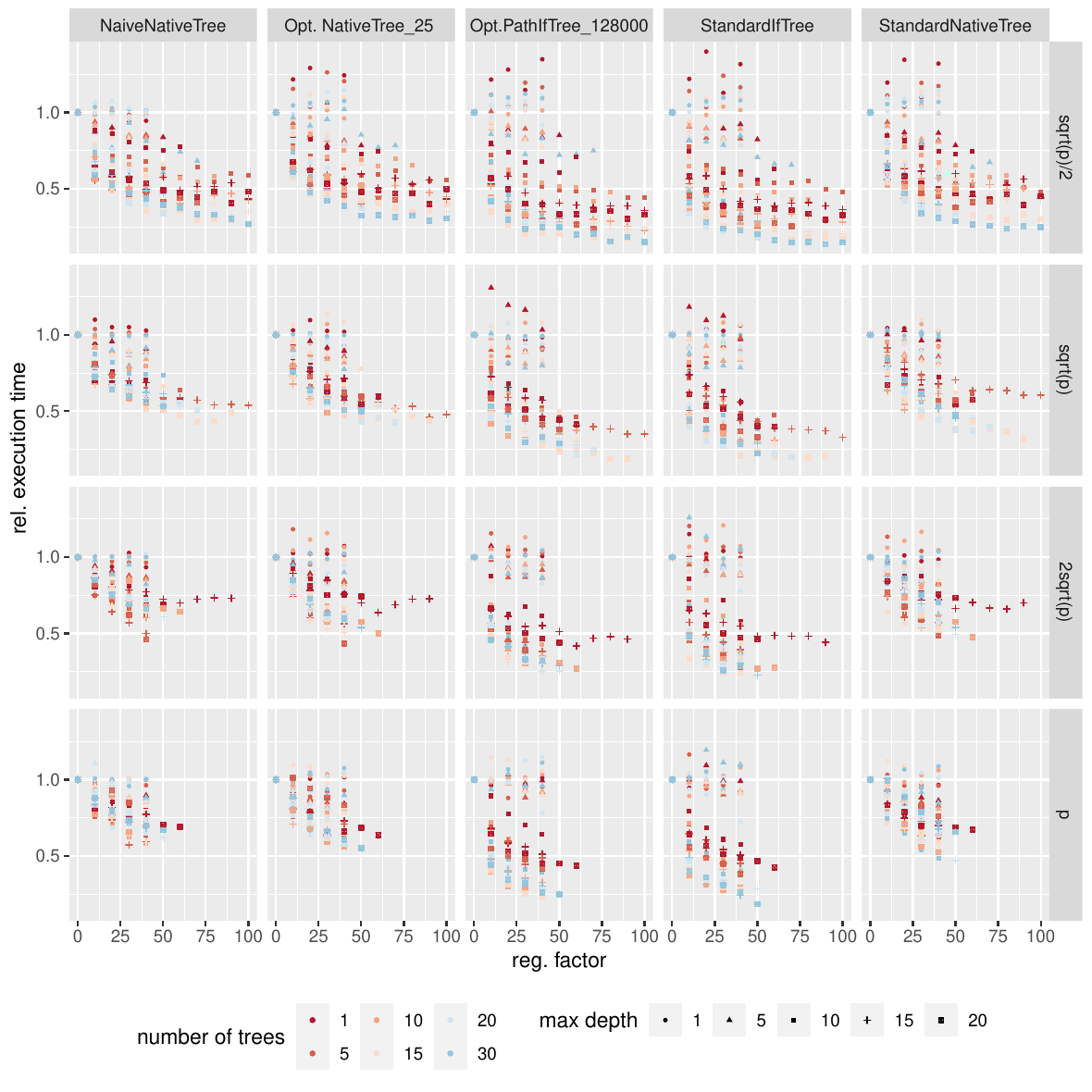} 
 \caption{Evaluation of the relative execution time for the magic dataset separated by the execution type $max\_features$, the maximum depth (shape of the points) and the number of trees (color).}
\end{figure*}
\begin{figure*}[ht]
    \centering
 \includegraphics[width=1\textwidth]{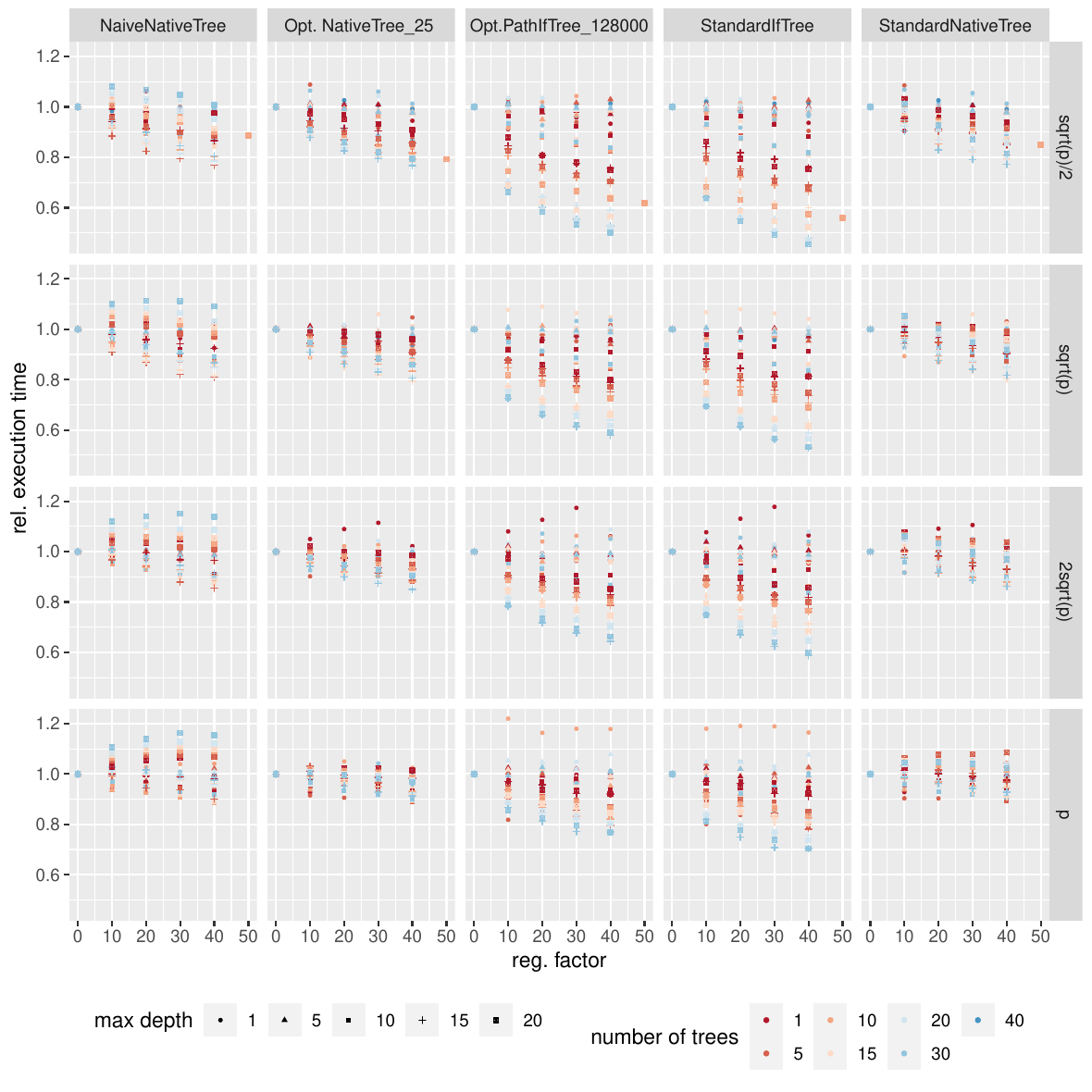} 
 \caption{Evaluation of the relative execution time for the mnist dataset separated by the execution type $max\_features$, the maximum depth (shape of the points) and the number of trees (color).}
\end{figure*}
\begin{figure*}[ht]
    \centering
 \includegraphics[width=1\textwidth]{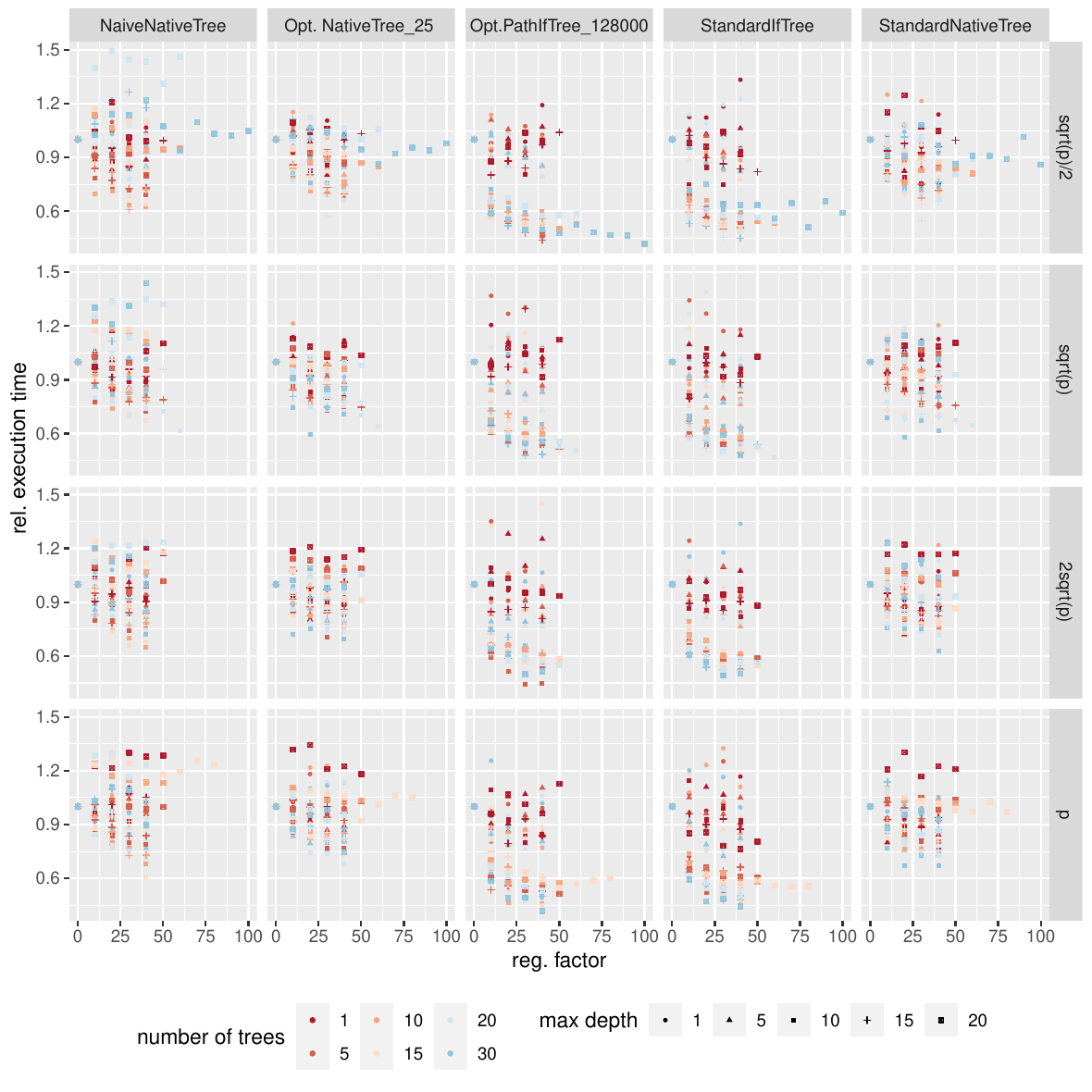} 
 \caption{Evaluation of the relative execution time for the satlog dataset separated by the execution type $max\_features$, the maximum depth (shape of the points) and the number of trees (color).}
\end{figure*}\begin{figure*}[ht]
    \centering
 \includegraphics[width=1\textwidth]{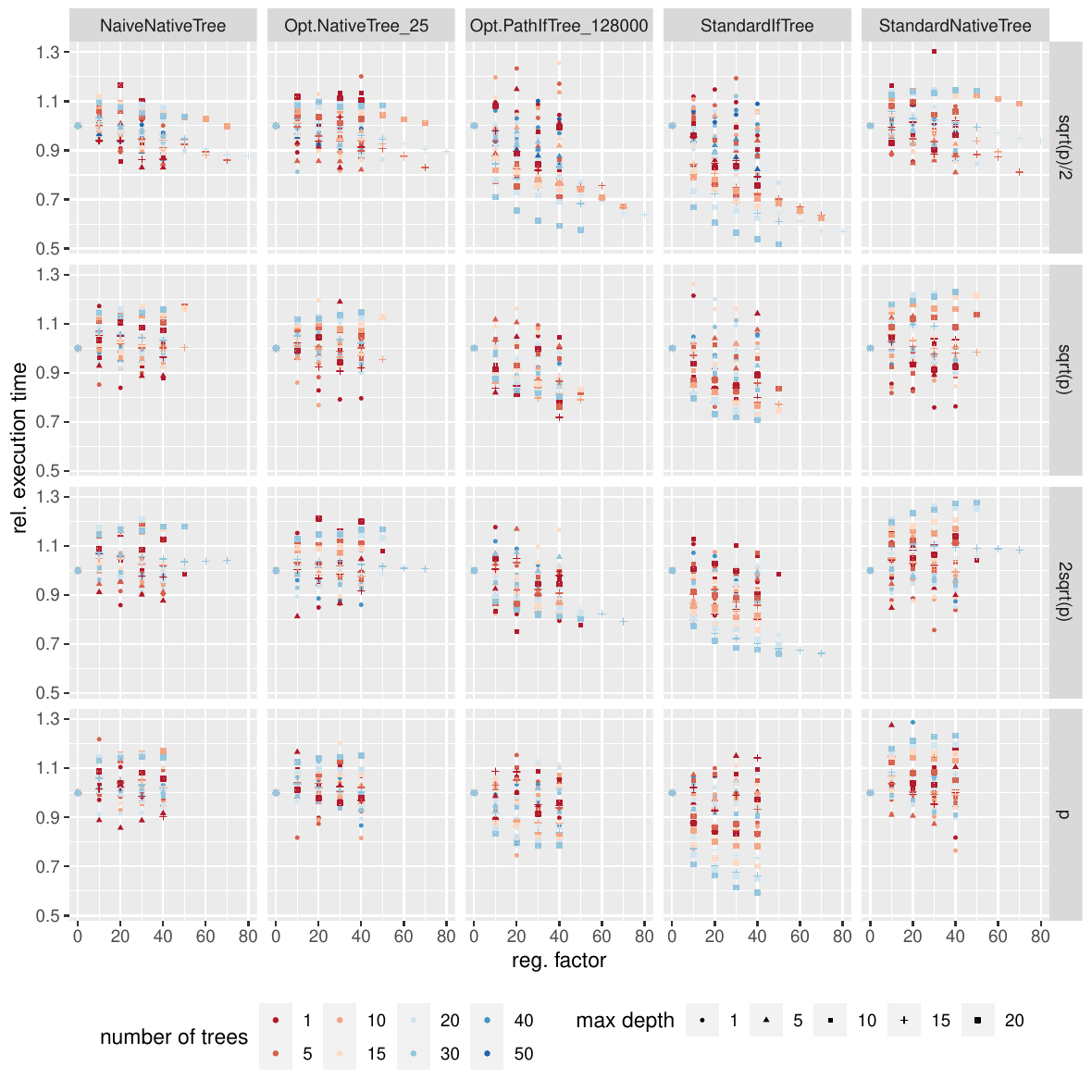} 
 \caption{Evaluation of the relative execution time for the sensorless dataset separated by the execution type $max\_features$, the maximum depth (shape of the points) and the number of trees (color).}
\end{figure*}\begin{figure*}[ht]
    \centering
 \includegraphics[width=1\textwidth]{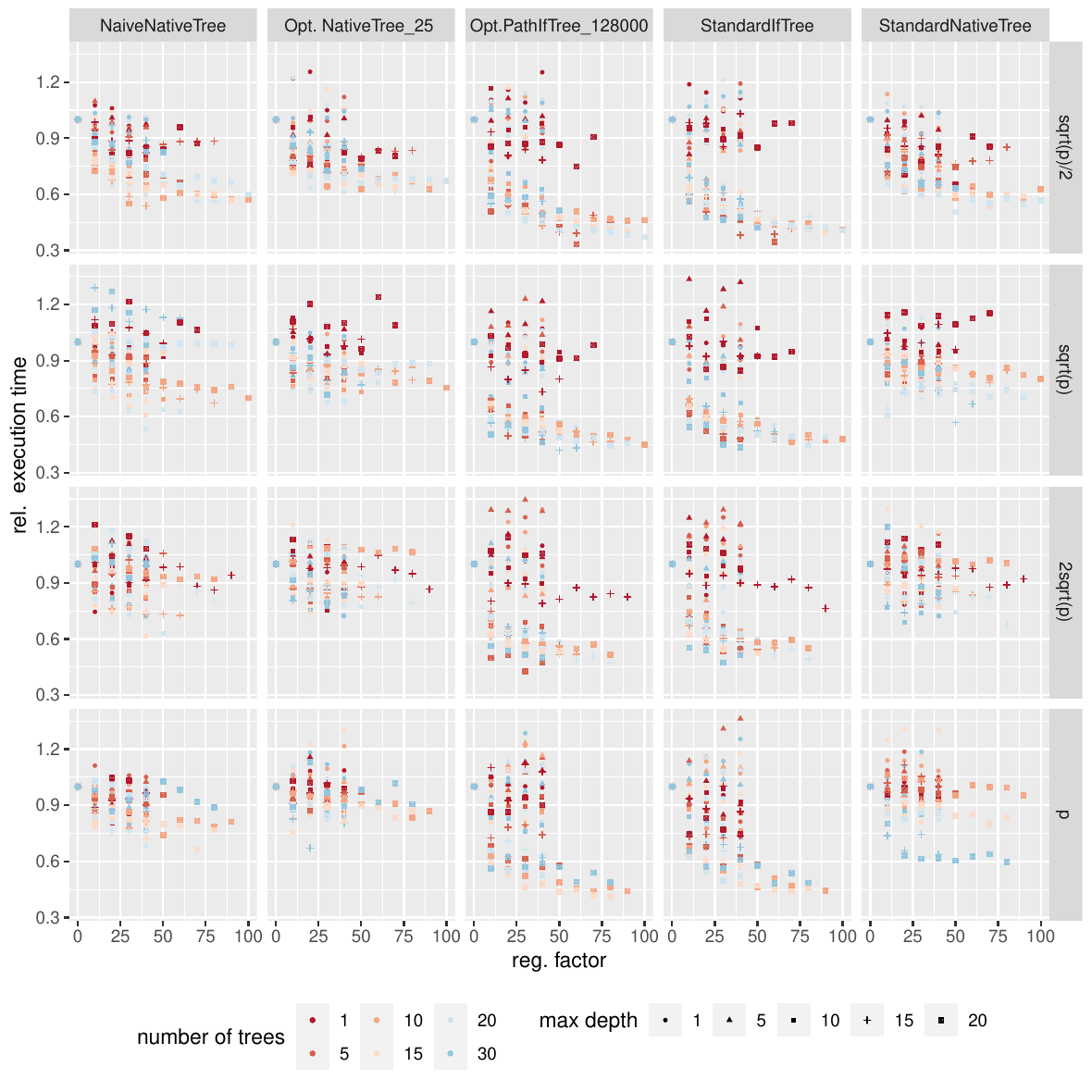} 
 \caption{Evaluation of the relative execution time for the spambase dataset separated by the execution type $max\_features$, the maximum depth (shape of the points) and the number of trees (color).}
\end{figure*}\begin{figure*}[ht]
    \centering
 \includegraphics[width=1\textwidth]{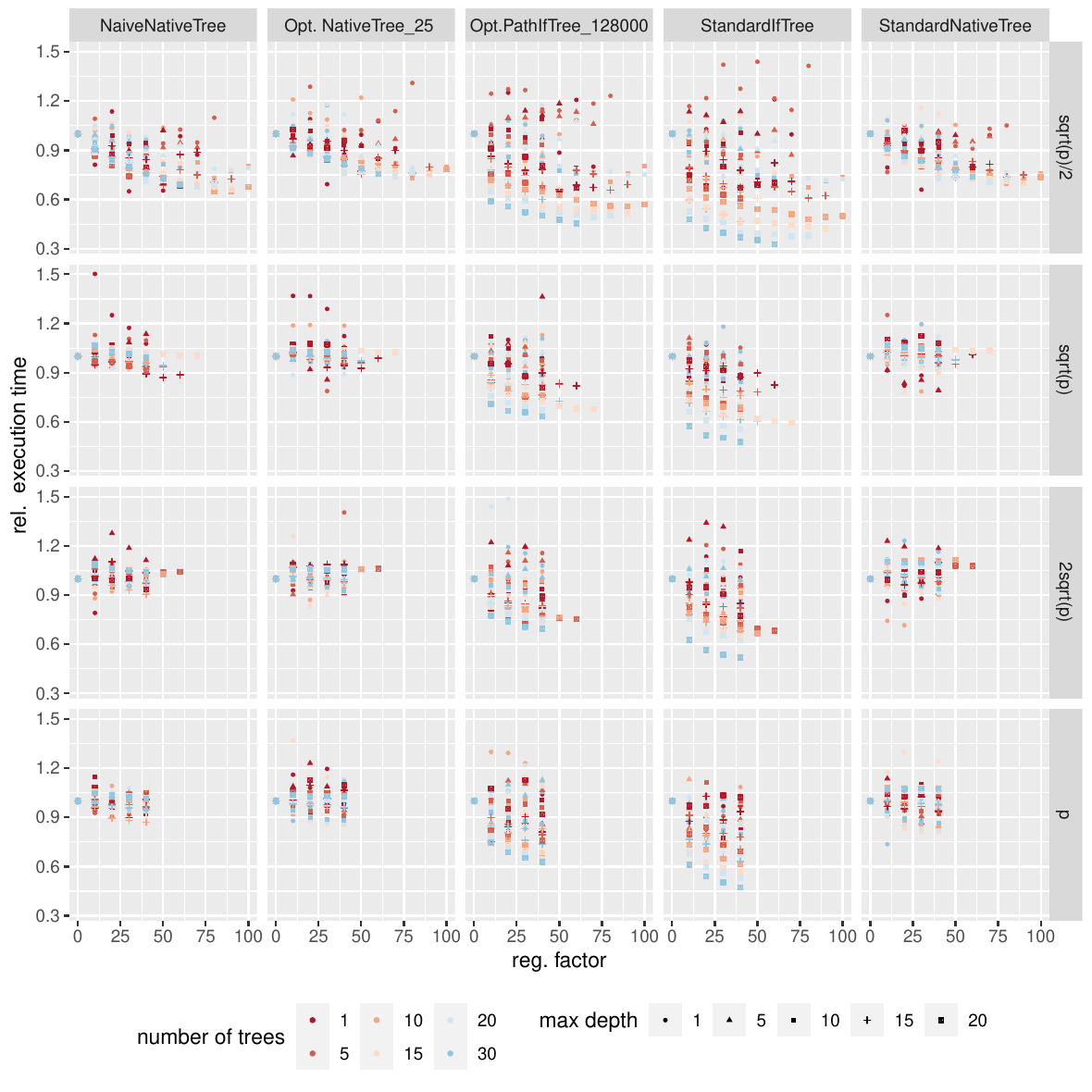} 
 \caption{Evaluation of the relative execution time for the wearable dataset separated by the execution type $max\_features$, the maximum depth (shape of the points) and the number of trees (color).}
\end{figure*}\begin{figure*}[ht]
    \centering
 \includegraphics[width=1\textwidth]{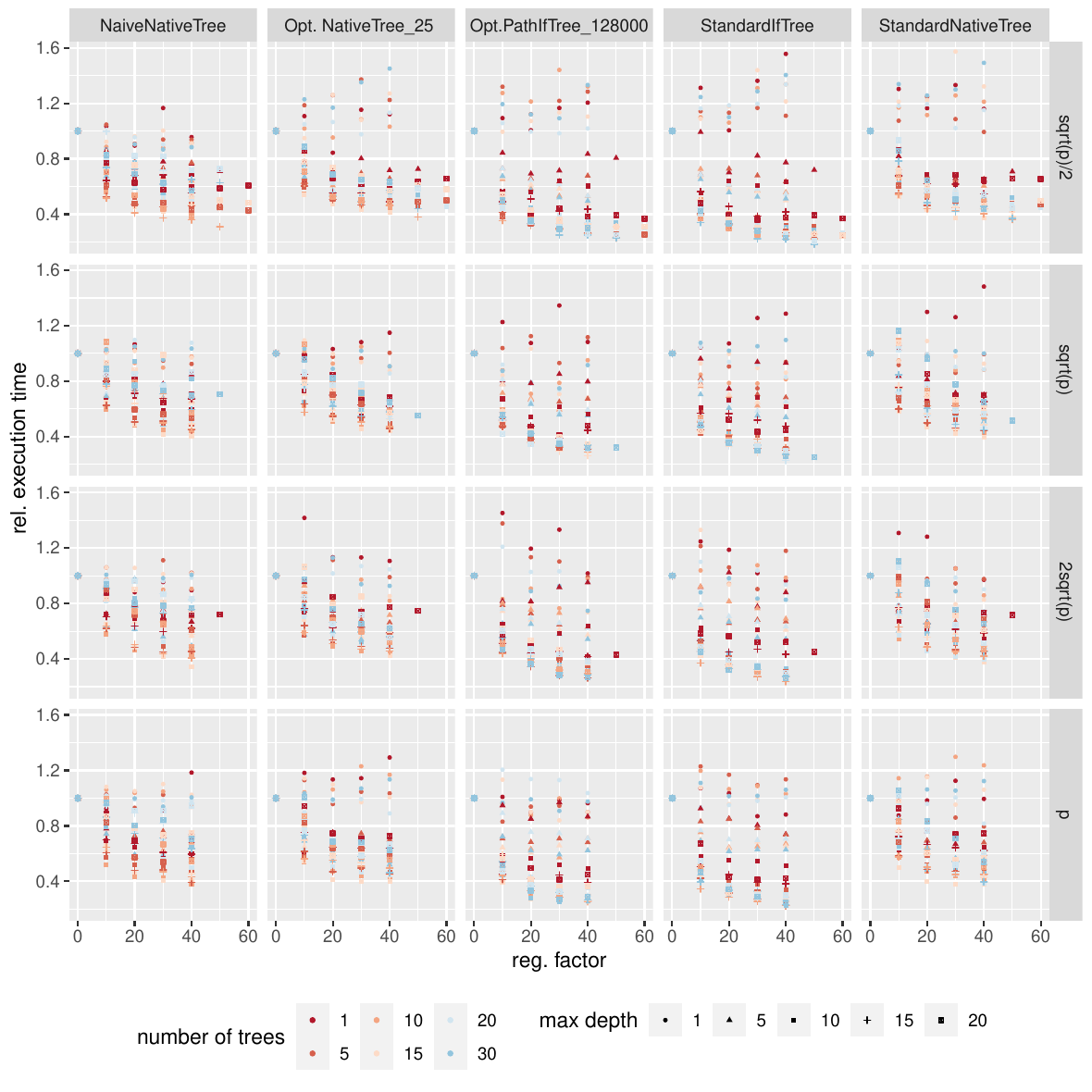} 
 \caption{Evaluation of the relative execution time for the wine-quality dataset separated by the execution type $max\_features$, the maximum depth (shape of the points) and the number of trees (color).}
\end{figure*}

\begin{figure*}
    \centering
    \includegraphics[width=1\textwidth]{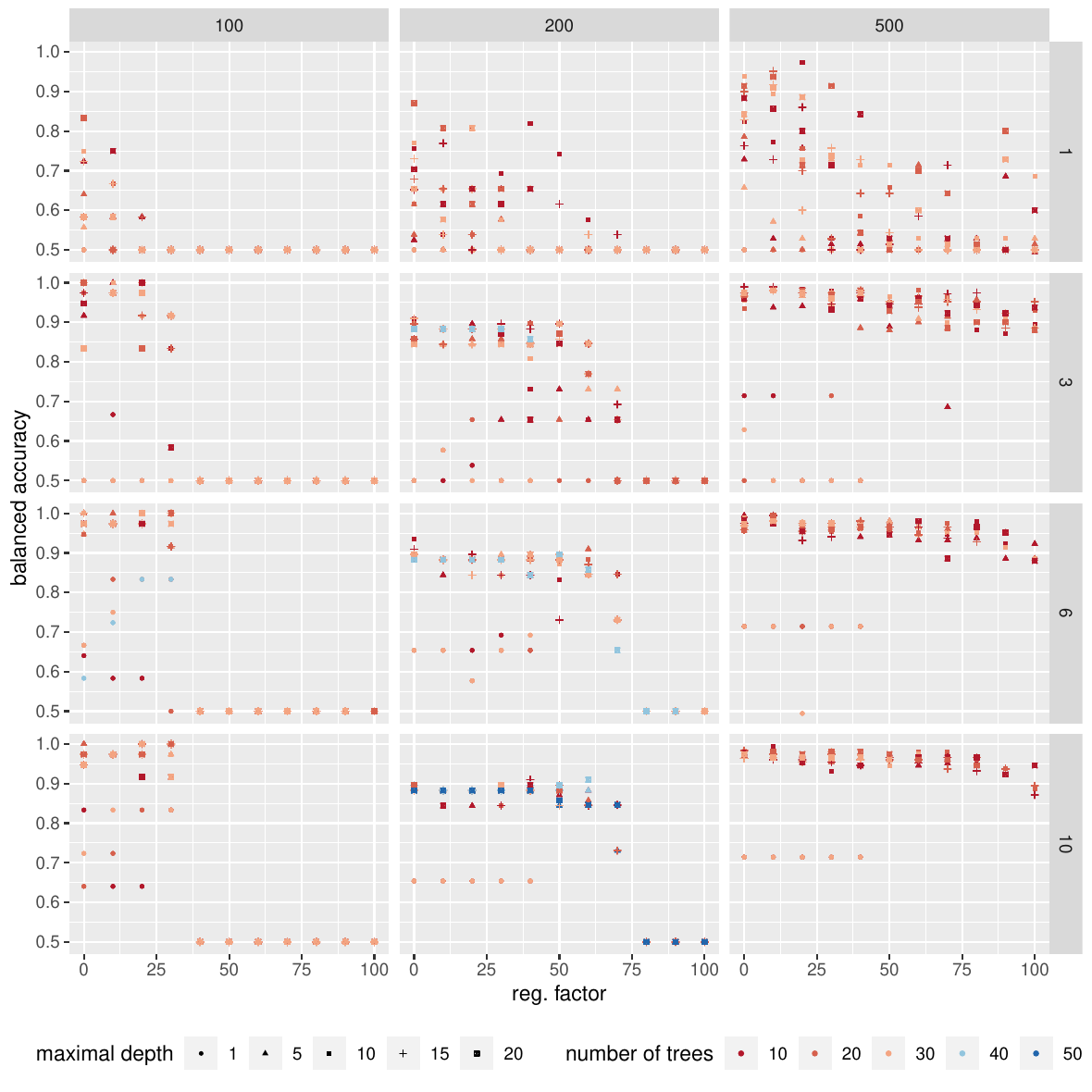}
    \caption{Balanced accuracy evaluation for synthetic data (red setting) with varying sample size, $max\_features$, maximum depth (point shape) and number of trees (color).}
\end{figure*}

\begin{figure*}
    \centering
    \includegraphics[width=1\textwidth]{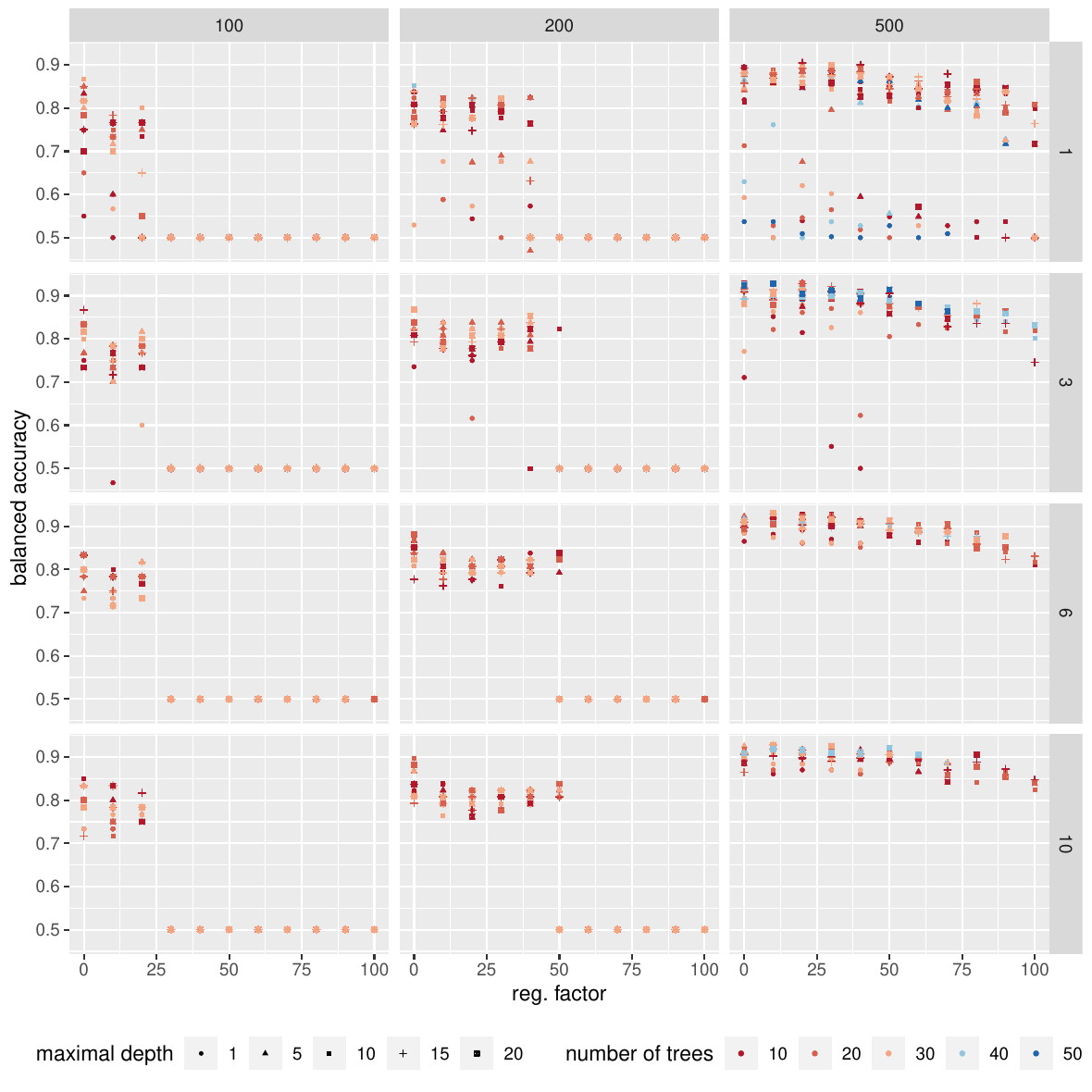}
    \caption{Balanced accuracy evaluation for synthetic data (green setting) with varying sample size, $max\_features$, maximum depth (point shape) and number of trees (color).}
\end{figure*}

\begin{figure*}
    \centering
    \includegraphics[width=1\textwidth]{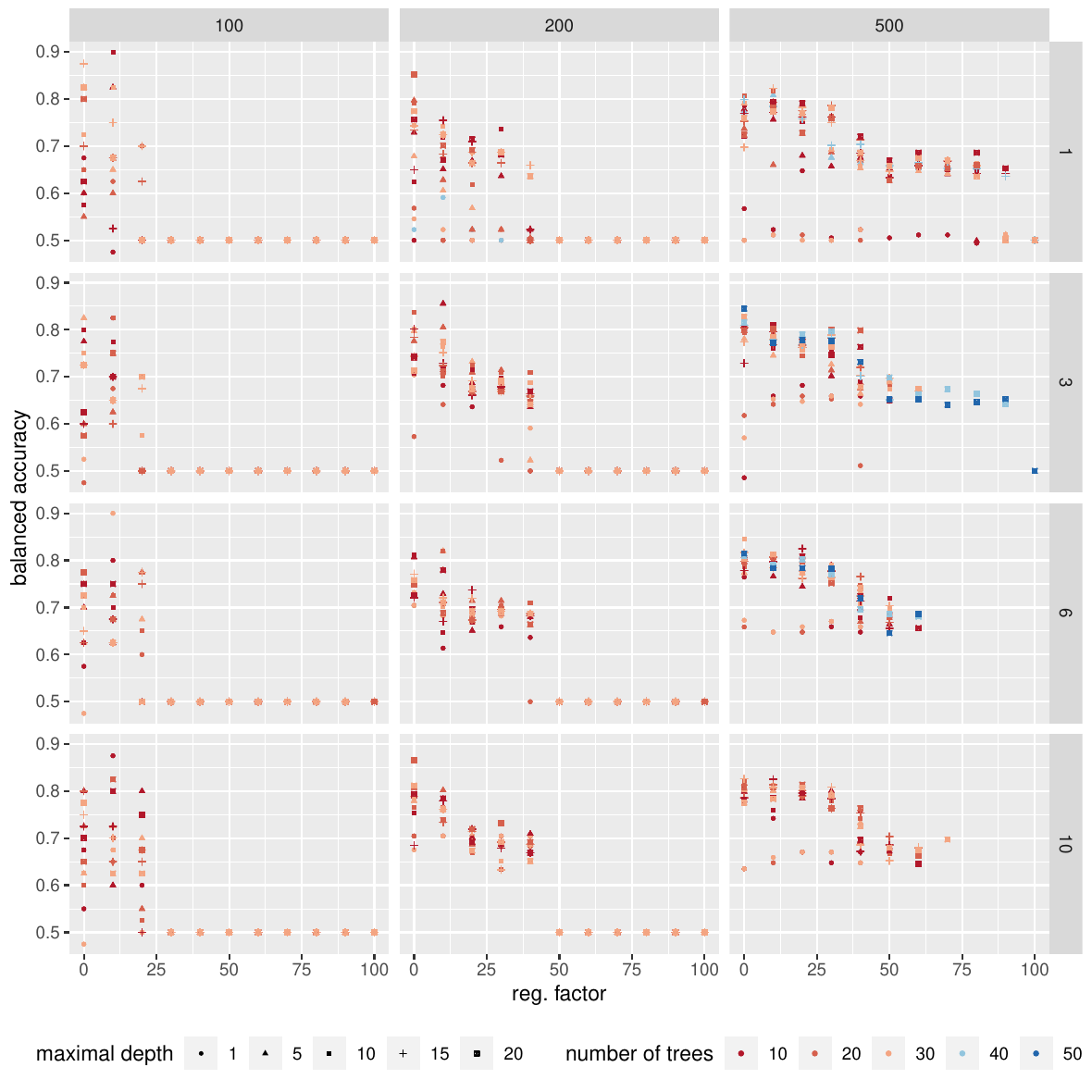}
    \caption{Balanced accuracy evaluation for synthetic data (blue setting) with varying sample size, $max\_features$, maximum depth (point shape) and number of trees (color).}
\end{figure*}

\begin{figure*}
    \centering
    \includegraphics[width=1\textwidth]{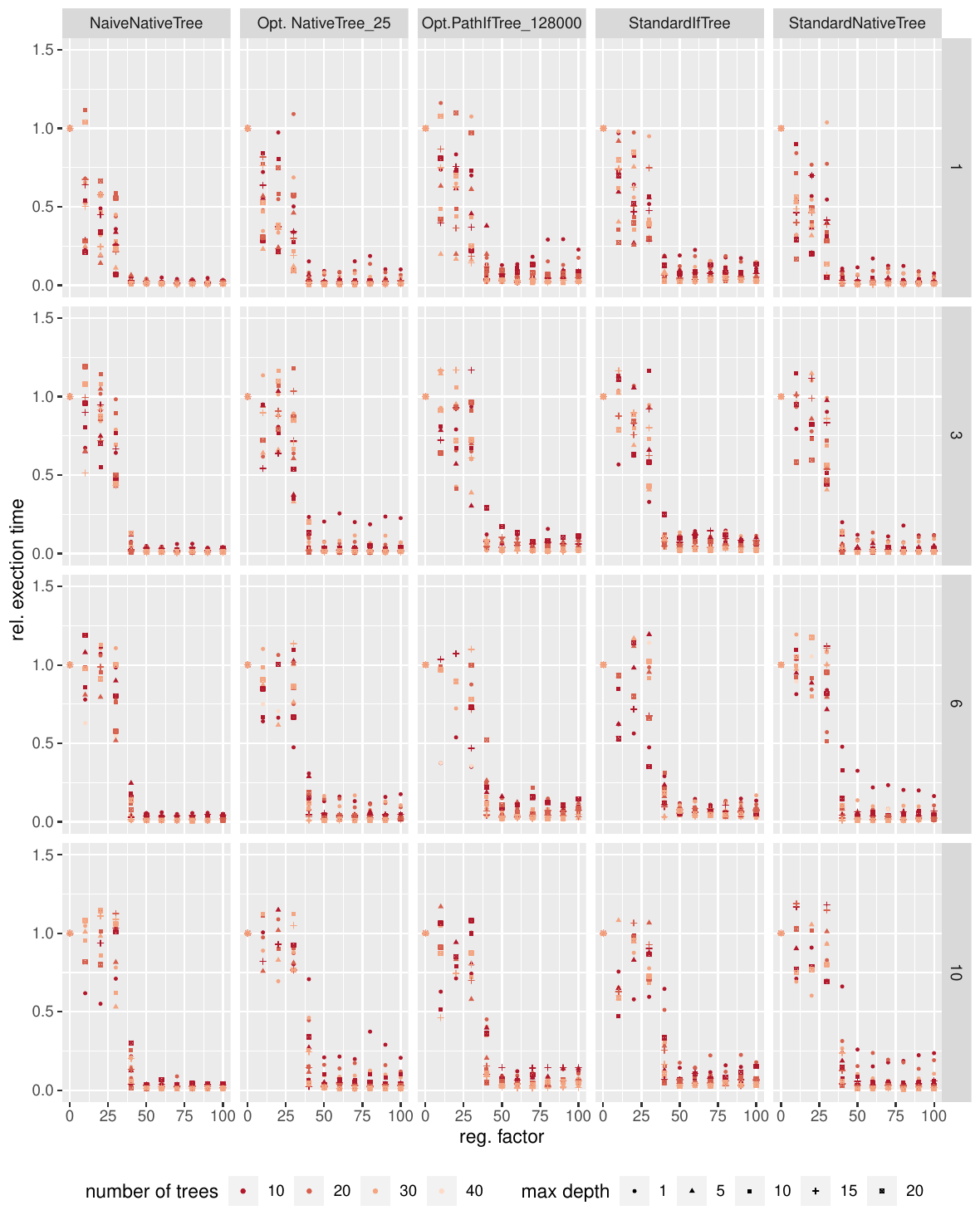}
    \caption{Relative execution time evaluation for synthetic data (red setting) and sample size=100 with varying sample size, $max\_features$, maximum depth (point shape) and number of trees (color).}
\end{figure*}

\begin{figure*}
    \centering
    \includegraphics[width=1\textwidth]{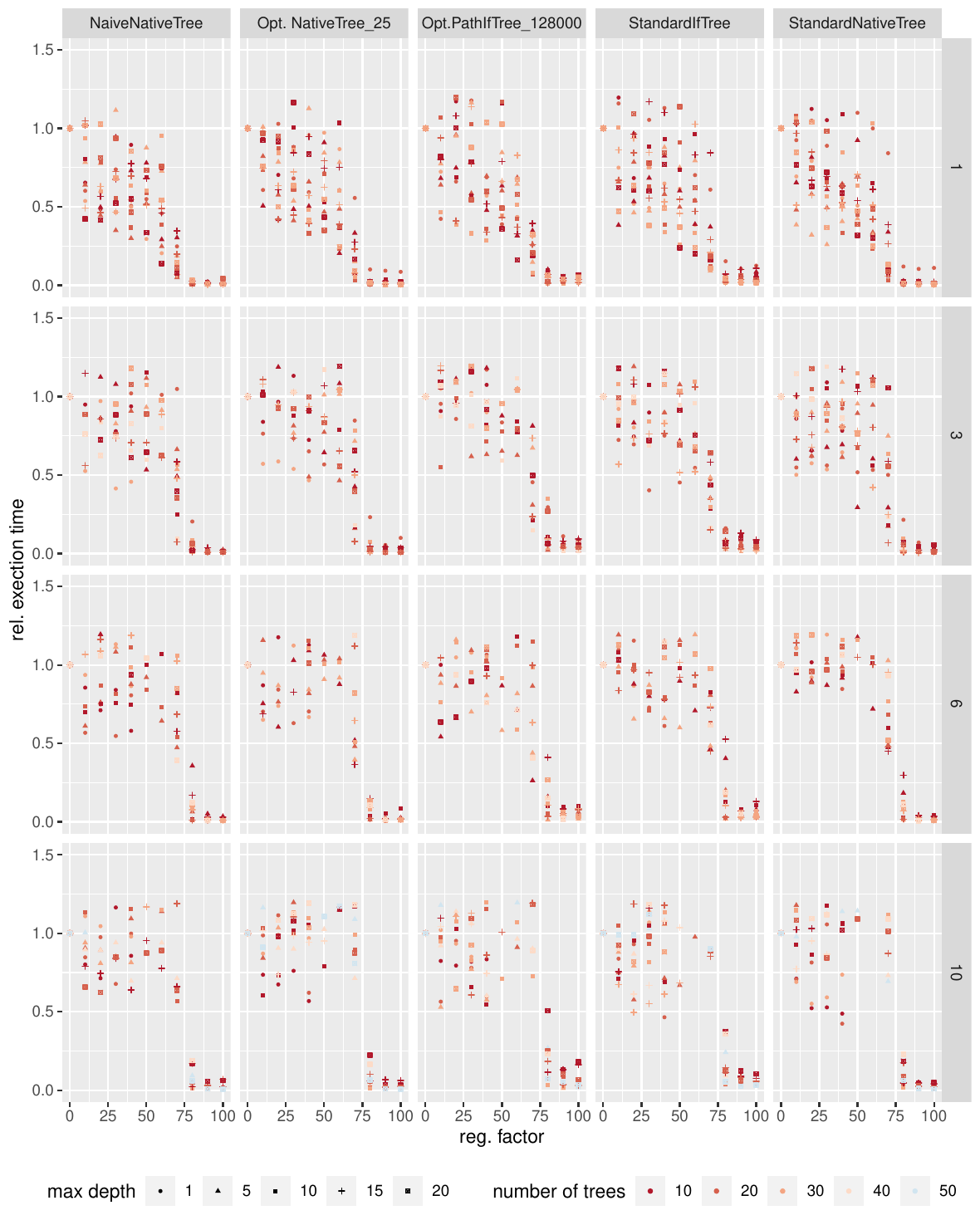}
    \caption{Relative execution time evaluation for synthetic data (red setting) and sample size=200 with varying sample size, $max\_features$, maximum depth (point shape) and number of trees (color).}
\end{figure*}

\begin{figure*}
    \centering
    \includegraphics[width=1\textwidth]{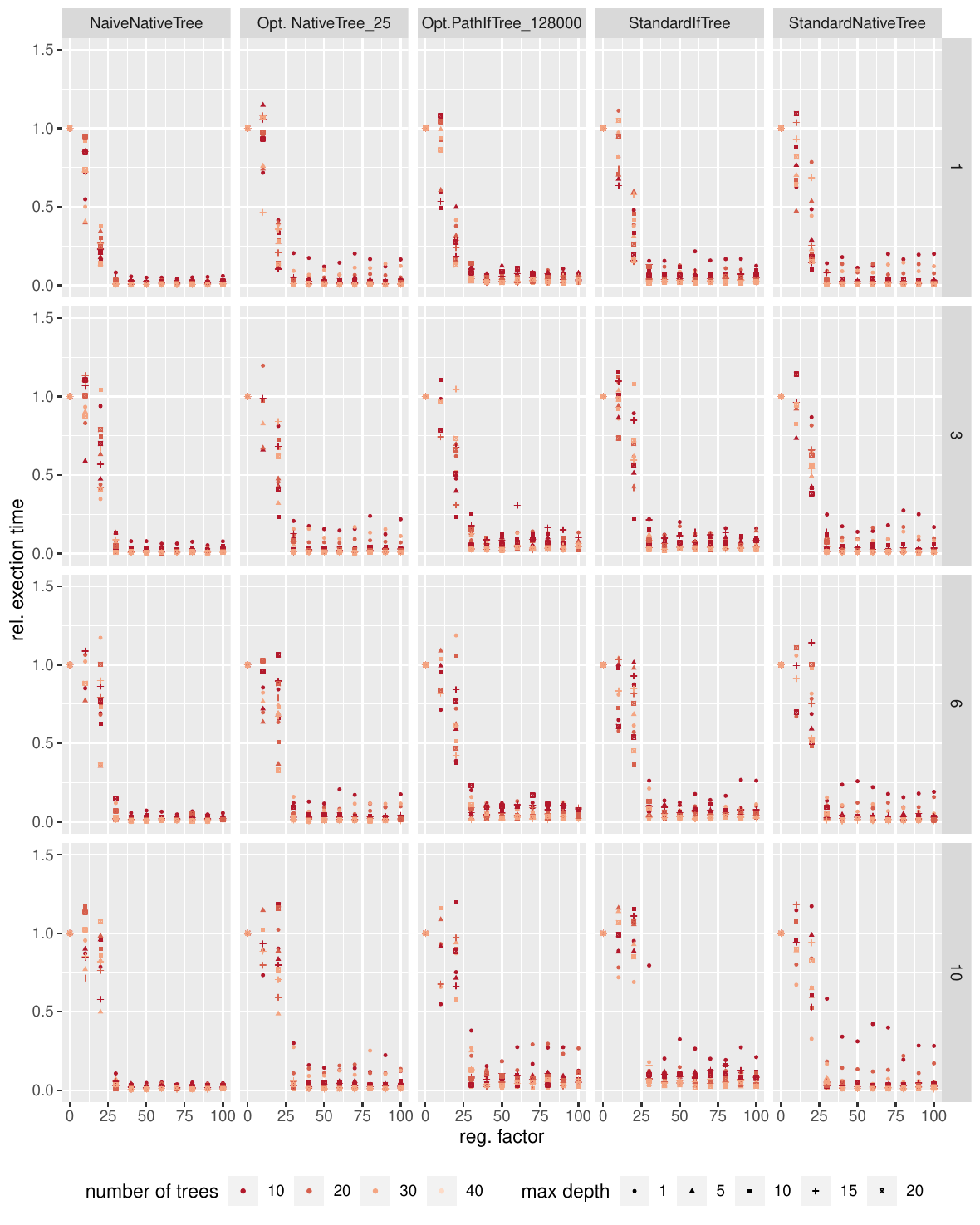}
    \caption{Relative execution time evaluation for synthetic data (green setting) and sample size=100 with varying sample size, $max\_features$, maximum depth (point shape) and number of trees (color).}
\end{figure*}

\begin{figure*}
    \centering
    \includegraphics[width=1\textwidth]{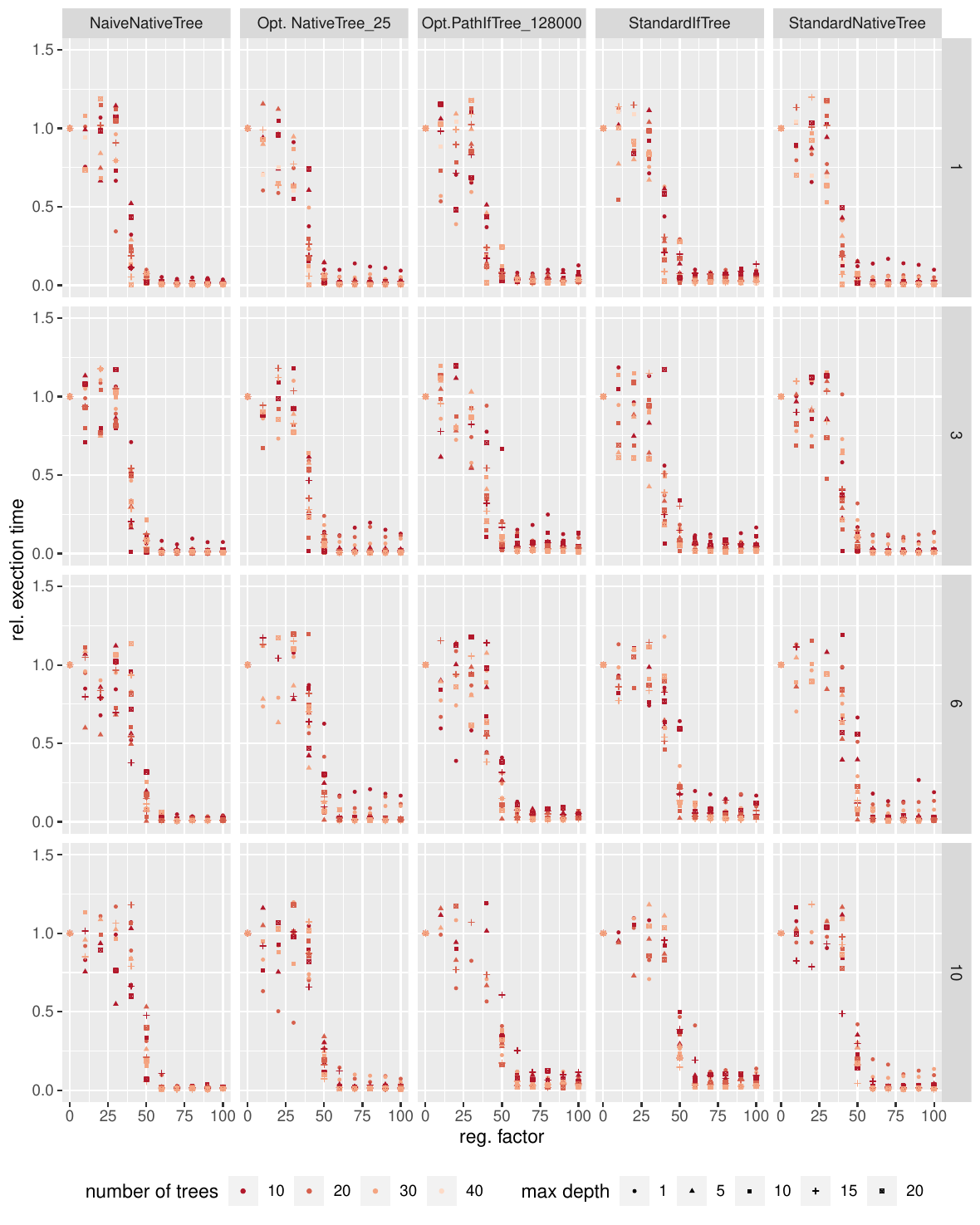}
    \caption{Relative execution time evaluation for synthetic data (green setting) and sample size=200 with varying sample size, $max\_features$, maximum depth (point shape) and number of trees (color).}
\end{figure*}

\begin{figure*}
    \centering
    \includegraphics[width=1\textwidth]{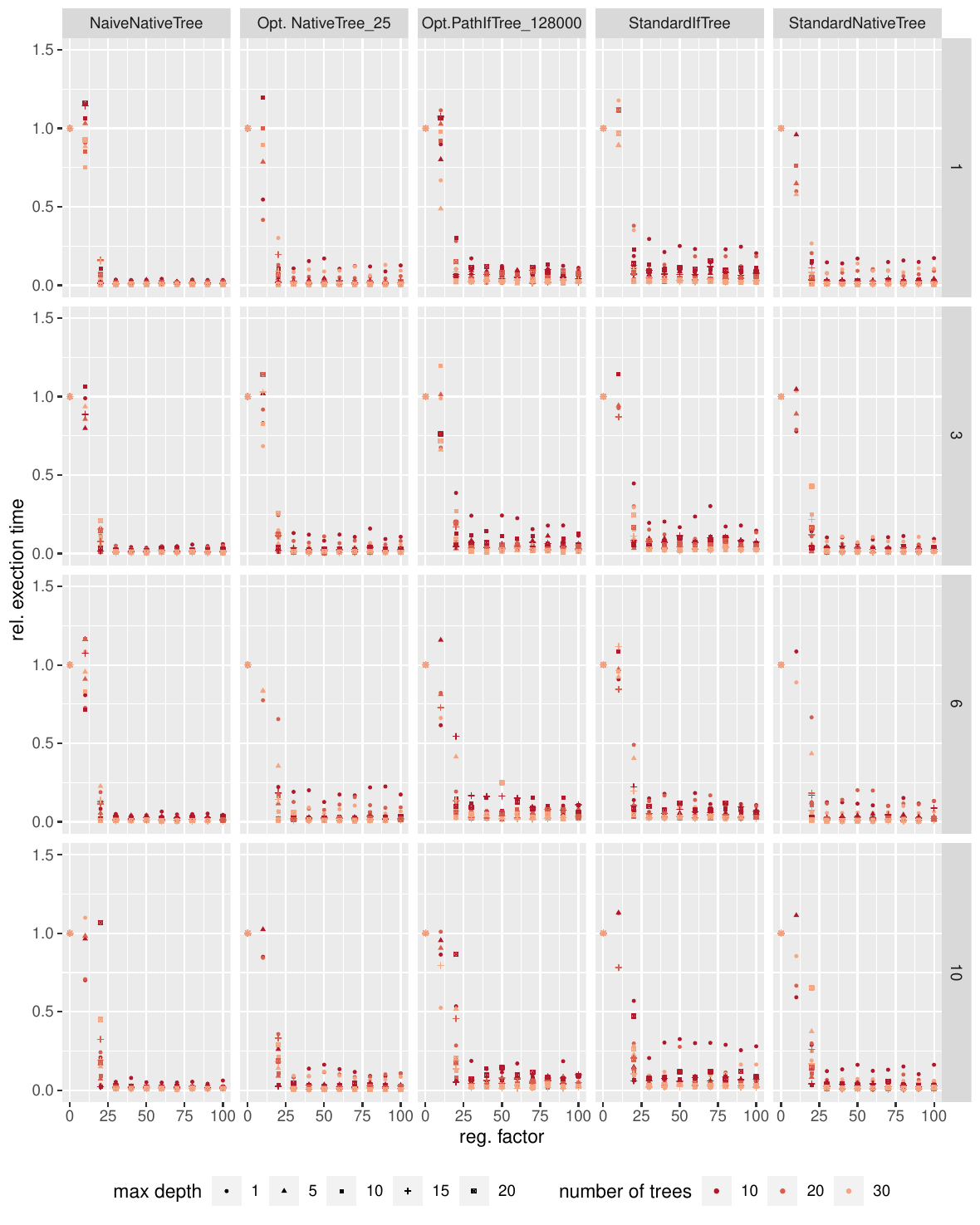}
    \caption{Relative execution time evaluation for synthetic data (blue setting) and sample size=100 with varying sample size, $max\_features$, maximum depth (point shape) and number of trees (color).}
\end{figure*}

\begin{figure*}
    \centering
    \includegraphics[width=1\textwidth]{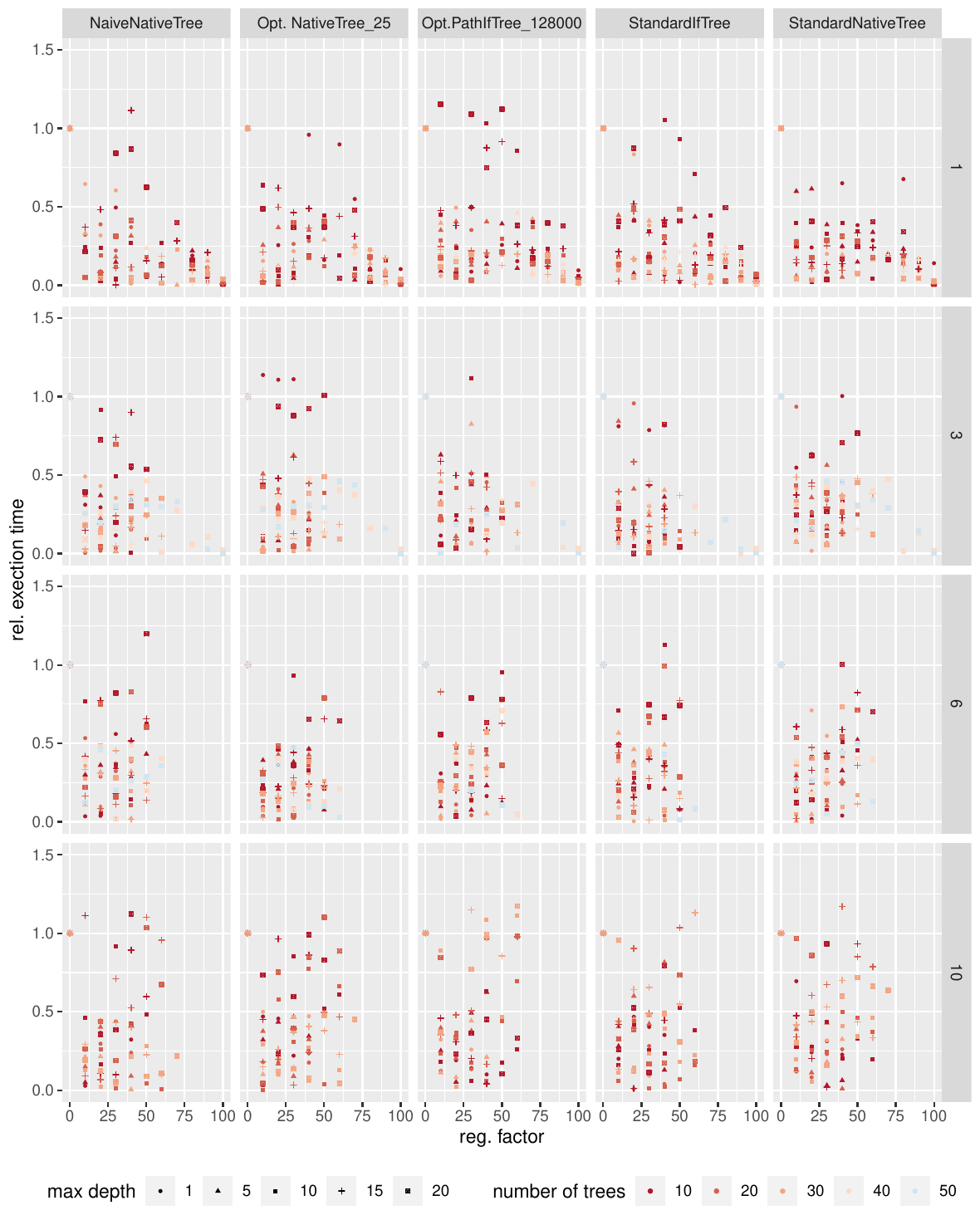}
    \caption{Relative execution time evaluation for synthetic data (blue setting) and sample size=200 with varying sample size, $max\_features$, maximum depth (point shape) and number of trees (color).}
\end{figure*}

\end{document}